\documentclass[10pt,twocolumn,letterpaper]{article}

\usepackage[pagenumbers]{cvpr} % To force page numbers, e.g. for an arXiv version

\usepackage{xspace}
\newcommand{\method}{AdaWorldPolicy\xspace}

\usepackage{multirow}
\usepackage{multicol}
\usepackage{adjustbox}
\usepackage{tabularray}
\usepackage{xcolor}
\usepackage{booktabs}
\usepackage{amssymb}

\usepackage{graphicx}
\usepackage{subcaption}
\usepackage{caption}
\usepackage{stfloats}
\usepackage{capt-of}
\usepackage{cuted} % 提供 strip 环境
\usepackage{xcolor}
\usepackage{newunicodechar}
\usepackage{afterpage}

\definecolor{DeepBlue}{RGB}{30, 60, 150}      % 更高饱和度深蓝色
\definecolor{VibrantGreen}{RGB}{20, 100, 20}   % 更高饱和度深薄荷绿
\definecolor{VibrantGreen2}{RGB}{50, 170, 50}   % 更高饱和度深薄荷绿
\definecolor{BrightRose}{RGB}{255, 60, 90}     % 更高饱和度深玫瑰粉
\definecolor{AmberYellow}{RGB}{255, 100, 0}    % 更高饱和度深金黄色

\definecolor{CrispRed}{RGB}{178, 34, 34}       % 火砖色，适合警告或强调
\definecolor{LivelyPurple}{RGB}{150, 114, 222}  % 中等紫色，适合辅助性说明

\definecolor{AlertRed}{RGB}{178, 34, 34}       % 红色，警示性强
\definecolor{RoyalPurple}{RGB}{148, 0, 211}    % 紫色，高贵且有深度

\definecolor{cvprblue}{rgb}{0.21,0.49,0.74}
\usepackage[pagebackref,breaklinks,colorlinks,allcolors=cvprblue]{hyperref}

\def\paperID{XXXX} % *** Enter the Paper ID here
\def\confName{CVPR}
\def\confYear{2026}

\title{\method: World-Model-Driven Diffusion Policy with Online Adaptive Learning for Robotic Manipulation}

\author{
    Ge Yuan$^{1}$, Qiyuan Qiao$^{2}$, Jing Zhang$^{2}$, Dong Xu$^{1}$\thanks{Corresponding author} \\
    {\tt\small gavinyuan97@gmail.com, qiaoqy@connect.hku.hk, zhang\_jing@buaa.edu.cn, dongxu@hku.hk} \\
    $^{1}$The University of Hong Kong
    $^{2}$Beihang University 
    \\
}

\begin{document}
\maketitle

%%% Choice 1. As teaser %%%
% \input{fig/fig1_teaser.tex}

\begin{abstract}
Effective robotic manipulation requires policies that can anticipate physical outcomes and adapt to real-world environments.
In this work, we introduce a unified framework, World-Model-Driven Diffusion Policy with Online Adaptive Learning (\textbf{\method}) to enhance robotic manipulation under dynamic conditions with minimal human involvement. Our core insight is that world models provide strong supervision signals, enabling online adaptive learning in dynamic environments, which can be complemented by force-torque feedback to mitigate dynamic force shifts.  
Our \method integrates a world model, an action expert, and a force predictor—all implemented as interconnected Flow Matching Diffusion Transformers (DiT). They are interconnected via the multi-modal self-attention layers, enabling deep feature exchange for joint learning while preserving their distinct modularity characteristics. We further propose a novel \textbf{O}nline \textbf{A}daptive \textbf{L}earning (AdaOL) strategy that dynamically switches between an Action Generation mode and a Future Imagination mode to drive reactive updates across all three modules. 
This creates a powerful closed-loop mechanism that adapts to both visual and physical domain shifts with minimal overhead.
Across a suite of simulated and real-robot benchmarks, our \method achieves state-of-the-art performance, with dynamical adaptive capacity to out-of-distribution scenarios.
Homepage: \url{https://AdaWorldPolicy.github.io}
\end{abstract}    
\section{Introduction}
\label{sec:intro}

%%% Coice 2. As single-columen %%%
% %%% Choice 1. As teaser %%%
% \begin{strip}
%   \centering
%   \includegraphics[width=\textwidth]{fig/awp-1_teaser.png}
%   \captionof{figure}{{An overview of our \textbf{\method} with Adaptive Online Learning (AdaOL).}
% Our framework operates in two synergistic modes. 
% {Mode I (Action Generation, center-left)}: The policy $\pi(a|o)$ takes the current multi-modal observation (static/gripper cameras and force sensor) and generates an action, which is then executed by the robot.
% {Mode II (Future Imagination, top-center)}: Conditioned on the executed action, the world model $\pi(o'|o, a)$ predicts the future imagined observation based on the same current observation. 
% The discrepancy between the \textit{Imagined Observation} and the \textit{Future Observation} (e.g., whether in in-domain setup or under domain shifts like lighting or pose variations) is used by our \textbf{AdaOL} strategy in a flow matching loss to compute a corrective gradient. This gradient is then used for updating an {Online Update} on a small subset of model parameters via LoRA in an online fashion, creating a closed-loop system that continuously adapts to real-world dynamic environments.
% \end{strip}

%%% Choice 2. As single-column %%%
\begin{figure}[t]
  \centering
  \includegraphics[width=\linewidth]{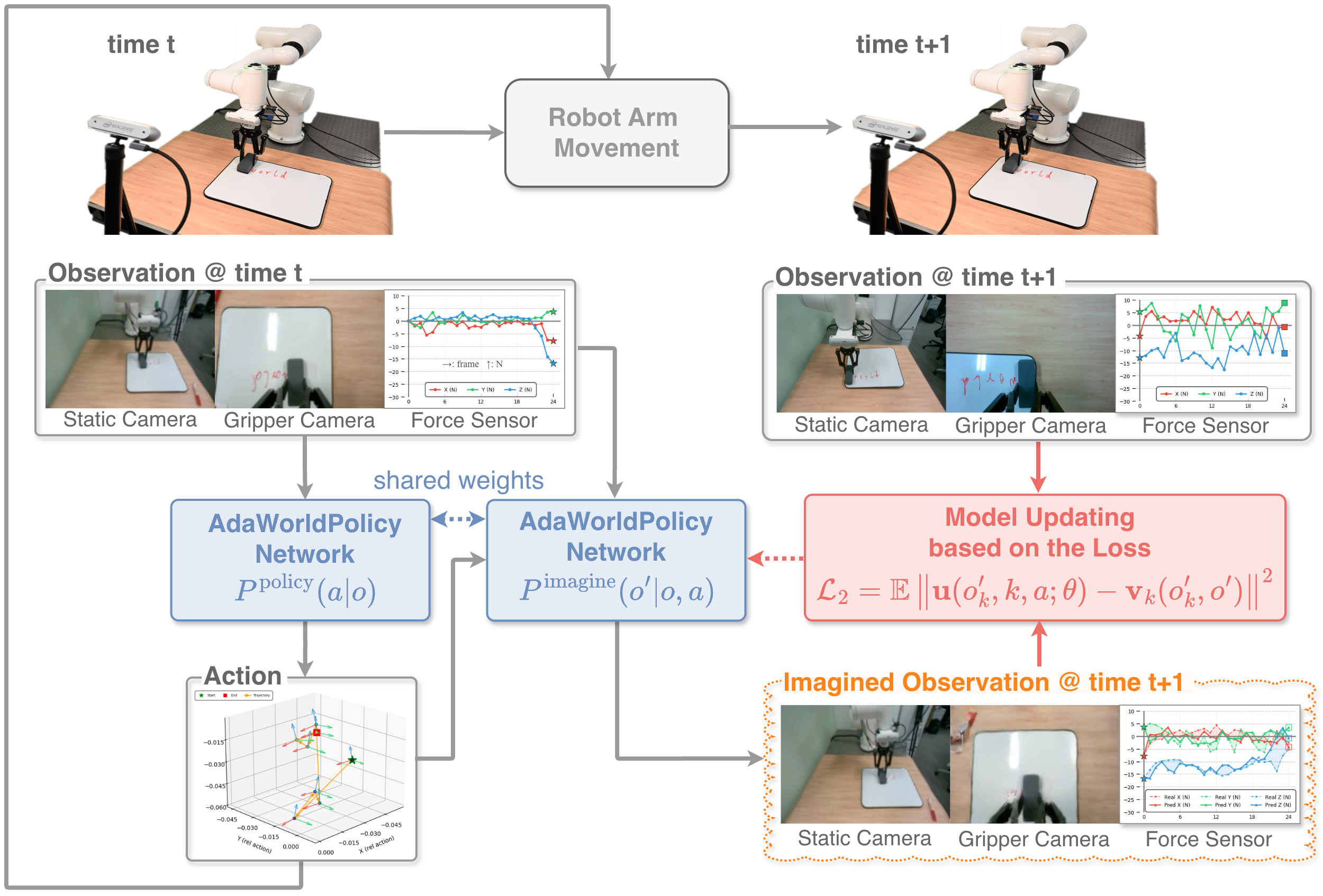}
    \captionof{figure}{{An overview of our \textbf{\method} with Adaptive Online Learning (AdaOL).}
    At timestep $t$, our {AdaWorldPolicy Network} operates in two modes. 
    \textbf{Mode I (Action Generation)}: AdaWorldPolicy network acts as an action policy generator $P^{\text{policy}}(a|o)$, which takes the current multi-modal observation $o$ (from static/gripper cameras and force sensor) to generate an action $a$. This action is then executed by the robot. During offline training, this step is supervised by the imitation loss $\mathcal{L}_1$ (see Section~\ref{sec:model_and_offline_training}).
    \textbf{Mode II (Future Imagination)}: Subsequently, our AdaWorldPolicy network turns into an action-conditioned world model $P^{\text{imagine}}(o'|o, a)$ which takes the same observation $o$ and the executed action $a$ to predict an \textit{Imagined Observation} at timestep $t+1$.
    The core of our \textbf{AdaOL} strategy lies in the online updating loop (red arrows). The discrepancy between the \textit{Imagined Observation} and the real \textit{Observation} at timestep $t+1  $ (e.g., under in-domain setup or under domain shifts like lighting or pose variations) is quantified by a prediction loss $\mathcal{L}_2$. This loss drives an online update to a small subset of shared network parameters, creating a closed-loop system that continuously adapts to real-world dynamics.
    % Our framework operates in two synergistic modes. 
    % {Mode I (Action Generation, center-left)}: The policy $\pi(a|o)$ takes the current multi-modal observation (static/gripper cameras and force sensor) and generates an action, which is then executed by the robot.
    % {Mode II (Future Imagination, top-center)}: Conditioned on the executed action, the world model $\pi(o'|o, a)$ predicts the future imagined observation based on the same current observation. 
    % The discrepancy between the \textit{Imagined Observation} and the \textit{Future Observation} (e.g., whether in in-domain setup or under domain shifts like lighting or pose variations) is used by our \textbf{AdaOL} strategy in a flow matching loss to compute a corrective gradient. This gradient is then used for updating an {Online Update} on a small subset of model parameters via LoRA in an online fashion, creating a closed-loop system that continuously adapts to real-world dynamic environments.
    }
  \label{fig:teaser}
\end{figure}

Robotic manipulation requires policies that can perceive, anticipate, and act reliably under contact-rich and dynamically changing conditions. 
Vision-Language-Action (VLA) models~\cite{pi0_black2410pi0, rt2_brohan2023rt, openvla_kim2024openvla} attempt to meet this requirement by integrating robotic action into large-scale pretrained Multimodal Large Language Models~\cite{eai_llm_tan2024true, eai_llm_szot2023large, eai_llm_zheng2023steve}. However, they normally require large amounts of human demonstration data and struggle to generalize to unseen or dynamic environments in contact-rich tasks.
% While VLA models excel at understanding high-level intent, their low-level control often struggles with long-horizon tasks, distribution shifts, and contact-rich challenges in the real world~\cite{rdp_xue2025reactive}.
% Two core issues persist: (1) it requires vast amounts of human-operated robot data to bridge the semantic gap between language and action without explicit physical modeling~\cite{openx_rtx2023, vla_langactgap_zhou2023bridging}, and (2) open-loop statistical matching lacks a feedback mechanism for action refinement, hindering robust generalization to unseen or dynamic scenarios~\cite{evalpi0_wang2025pi0_wild, purevlasurvey_zhang2025pure}.
% Concurrently, due to the world simulation ability of world models~\cite{naviworldmodel_bar2025navigationWM, cosmos2_ali2025world, jepa2_assran2025vjepa2}, some methods~\cite{dinowm_zhou2024dino, enerverseac_jiang2025enerverse} directly use a world model as the policy validator to provide rich information about real-world dynamics and agent-environment interactions.
Concurrently, researchers have explored how to integrate world models~\cite{naviworldmodel_bar2025navigationWM, cosmos2_ali2025world, jepa2_assran2025vjepa2, cosmos1_agarwal2025cosmos} for robot manipulation by providing rich information about real-world dynamics and agent-environment interactions.
However, many existing approaches only use the world model as a passive ``digital twin" for offline evaluation~\cite{enerverseac_jiang2025enerverse, dinowm_zhou2024dino}. 
Another more integrated strategy attempts to create a unified architecture for both action generation and world prediction~\cite{uva_li2025unified, zhu2025unified, worldvla_cen2025worldvla}.
These methods typically adopt the offline training strategy that fails to dynamically adapt to real-world environmental changes. 
Therefore, how to fully exploit the world model's potential for robotic manipulation in dynamic scenarios with minimal human efforts remains a challenging task.

In this work, we enhance robot manipulation when facing environmental changes and dynamic force shifts in a reactive manner with minimal human involvement. Our key insight is that world models provide strong supervision signals for robotic manipulation and enable online adaptive learning in dynamic environments. Additionally, we find that force-torque feedback is valuable to reduce dynamic force shifts during real-world deployment on contact-rich tasks.

To this end, we introduce \method, a unified multi-modal framework consisting of three components: a world model, an action expert, and a force predictor. All three components are implemented as Flow Matching~\cite{flowmatching_lipman2022flow} Diffusion Transformers (DiT)~\cite{dit_peebles2023scalable}. The world model is built upon the state-of-the-art Cosmos Predict2~\cite{cosmos2_ali2025world}, while the action expert and force predictor are lightweight DiT models. These three modules operate in parallel and are interconnected via the multi-modal self-attention layers, enabling deep feature exchange while preserving their distinct computational pathways. The world model provides supervision signals for the action expert and also enables online adaptive learning in dynamic environments by identifying discrepancies between its action-conditioned predictions and actual real-world feedback. The force predictor addresses dynamic force shifts by minimizing the discrepancy between predicted and actual force readings. To jointly train the three modules, we propose a test-time online adaptive learning (AdaOL) strategy where \method switches between the following two modes. In \textit{Mode I (Action Generation)}, the action model generates actions based on the current observations. In \textit{Mode II (Future Imagination)}, two types of discrepancies including these between the world model's action-conditioned predictions and actual real-world feedback, as well as these between predicted and actual force readings, drive self-supervised online updates for the network parameters in all three modules. These modules are updated in a reactive manner to achieve fast adaptation and closed-loop control in response to dynamically changing environments.

Our main contributions are summarized as follows:
\begin{itemize}
    \item \textbf{A unified multi-modal framework, \method.} This framework fully exploits the potential of world models and force-torque feedback for robotic manipulation in contact-rich dynamic environments by synergistically modeling an action expert, the pre-trained world model, and the force predictor within a unified diffusion-based network architecture.
    % that synergistically models action expert, world models, and force predictor within a shared diffusion-based architecture, enabling deep feature fusion and efficient joint updates.
    \item \textbf{A novel test-time online adaptive learning (AdaOL) strategy}. The AdaOL strategy rapidly reduces both visual and physical domain shifts by updating the network parameters based on real-world feedback.
    % that creates a closed-loop control mechanism. It rapidly adapts to both visual and physical domain shifts by correcting the internal models of action and dynamics based on real-world feedback.
    % \item \textbf{A practical demonstration of modality extension} by integrating a force expert, showcasing our framework's flexibility to incorporate non-visual feedback crucial for contact-rich manipulation and adapt to shifts in physical dynamics.
    \item \textbf{State-of-the-art performance across diverse benchmarks.} Our base \method framework achieves state-of-the-art results on a suite of benchmarks, including PushT~\cite{diffusionplanning_janner2022planning}, CALVIN~\cite{calvin_mees2022calvin}, and LIBERO~\cite{libero_liu2024libero}. Furthermore, we demonstrate that our base AdaWorldPolicy with AdaOL improves out-of-distribution (OOD) performance by over 5\%, meanwhile  enhancing in-domain results by $\sim$1\%. Real-world experiments validate the effectiveness of both AdaOL and the force-aware extension in completing challenging manipulation tasks.
\end{itemize}

\section{Related Work}
\label{sec:related}

\paragraph{World Models for Robotic Control.}
Robotic policy learning methods are broadly divided into three paradigms. The first one is based on model-free, end-to-end visuomotor learning, where transformer-based models ingest raw visual (and sometimes language) input and directly output robot actions. For example, generalist systems are trained on vast teleoperation datasets~\cite{rt1_brohan2022rt, rt2_brohan2023rt, octo_team2024octo, openvla_kim2024openvla}. Although these agents perform well in multi-task scenarios through large-scale pre-training, they lack the capacity for explicit physical or dynamic reasoning, which degrades their performance in novel or out-of-distribution environments.

The second paradigm is based on model-based learning. Rather than mapping observations directly to actions, agents learn an internal world model that predicts environment dynamics, enabling planning, policy optimization, or latent imagination~\cite{hafner2020dreamer, hafner2025dreamerv3, schrittwieser2020mastering}. While these methods offer stronger physical grounding and better use of training data, traditionally they have been applied in narrower domains or simulated settings.

More recently, hybrid approaches are emerging: world models are either used as external validators over large pre-trained modules (e.g., large vision-language models)~\cite{dinowm_zhou2024dino}, or they are tightly integrated into unified architectures that combine world modeling, perception and policy in one system~\cite{uva_li2025unified, worldvla_cen2025worldvla}. In contrast, our work belongs to this hybrid paradigm: we propose a novel fusion mechanism in which the world model actively supervises policy learning and also corrects its output, rather than passively validating it.

\paragraph{Diffusion Models for Decision Making.}
In parallel, diffusion model-based approaches have been used for decision-making and control. Early works such as Diffusion Policy and its extensions model robot trajectories via diffusion processes~\cite{diffusionpolicy_chi2023diffusion, diffusionplanning_janner2022planning, ajay2022conditional}. These models outperform traditional behavior cloning methods in many multi-modal settings. However, a major drawback is that they often ignore dynamics and outcome modeling. Specifically, because they are trained via behavior cloning, they may generate actions that seem plausible but are physically inconsistent. More recent works aim to address this issue by introducing hierarchical generation strategies, kinematics-aware constraints, and contact-aware trajectory rollout~\cite{ma2024hierarchicalDiffusionPolicy, bouvier2025DDAT}. Following this research trend, we introduce a world model within the diffusion process to explicitly impose physical consistency while using model-predicted rollouts to generate better actions.

\paragraph{Online Adaptation for Robotics.}
Deploying robots in real-world, dynamic, and previously unseen environments is challenging due to distribution mismatch between training and testing processes. Test-time adaptation (TTA) provides a mechanism for agents to adjust after deployment based on unlabeled inputs. In vision and perception domains, TTA methods based on self-supervised objectives or memory buffers have shown promising results~\cite{TST_sun2020test, wang2022continual, adaptdiffuser_liang2023adaptdiffuser}. For large models, parameter-efficient fine-tuning (e.g., LoRA) enables rapid updates during inference with minimal overhead~\cite{lora_hu2022lora}. More recently, robotics-specific TTA approaches have been proposed, including plug-and-play transformer modules for adaptation~\cite{chang2024PLUTO}, low-rank adaptation based on confidence maximization~\cite{imam2025TTL}, and embodied adaptation for real robotic grasping or navigation tasks~\cite{liu2025embodiedTTA}. Our Adaptive Online Learning (AdaOL) strategy extends these works by using prediction error from the improved world model powered module as a self-supervised signal, enabling fast, physically grounded adaptation to both visual and dynamic force shifts.

In summary, compared to existing works, our \method{} introduces a unified diffusion-based framework where the world model, action expert, and force/contact predictor are deeply integrated through a multi-modal self-attention mechanism. Rather than simply validating or predicting, the world model becomes part of the learning loop—its prediction error serves as a self-supervised adaptation signal for correction of both visual and dynamic force shifts.
\section{Methodology}
\label{sec:method}

\begin{figure*}[ht]
    \centering
    \includegraphics[width=0.95\textwidth]{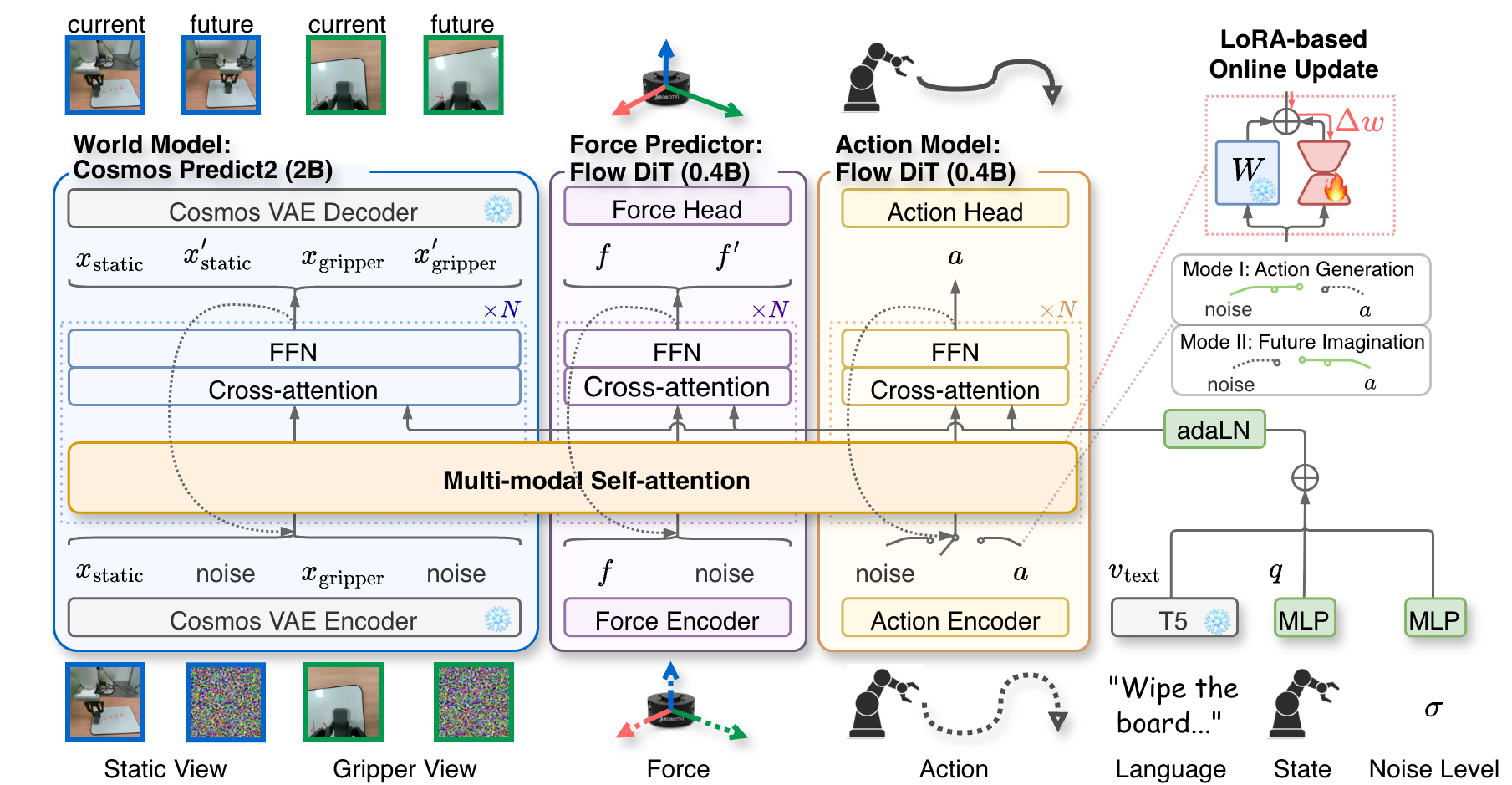}
    \caption{\textbf{Network architecture and workflow of \method.}
Our unified multi-modal framework builds upon a shared multi-modal transformer backbone. It synergistically integrates three modules: a \textbf{World Model} for visual prediction, a \textbf{Force Predictor} for physical dynamics modeling, and an \textbf{Action Model} for policy generation. All modules are implemented as Flow Matching Diffusion Transformers (DiT) and interact through a shared Multi-modal Self-attention layer. Input modalities (vision, action, force, text) are first encoded, conditioned with global features (text, state, noise level) via the adaLN module, and then processed by the shared multi-modal self-attention layer. In our framework, the operational mode is determined by a switch on the action input: in \textbf{Mode I (Action Generation)}, the action token is provided as noise for the model to generate an action; in \textbf{Mode II (Future Imagination)}, a known action is provided as a condition for future prediction. A LoRA-based mechanism enables efficient online updates of a small set of parameters.
}
    \label{fig:overall}
\end{figure*}

%-----------------------------------
\subsection{Problem Formulation}
\label{sec:problem_formulation}

We address the problem of learning a goal-conditioned robotic manipulation policy that can rapidly adapt to novel dynamics at test time without requiring new human demonstrations.
Formally, at each timestep $t$, the input is a multi-modal observation history 
% $o = \{o_t^1, o_t^2, f_t\}$, 
% $o = \{x_{\text{static},{t-T_\text{c}+1:t}}, o_{t-T_\text{c}+1:t}^2, f_{t-T_\text{c}+1:t}\}$,
$o = \{x_{\text{static}}, x_{\text{gripper}}, f\}$,
which includes image sequences from a static camera $x_{\text{static},t-T_\text{c}+1:t}$ and a gripper camera $x_{\text{gripper},t-T_\text{c}+1:t}$, and force-torque sensor readings $f_{t-T_\text{c}+1:t}$, all sharing the same context length $T_\text{c}$. The goal is to generate a sequence of future actions $a=a_{t:t+T_\text{a}-1}$ with horizon $T_\text{a}$ that successfully completes the task.

% The core challenge lies in the test-time adaptation requirement. The policy is initially trained offline on a static dataset collected in the source domain. However, when deployed in a new test environment with different dynamics, the initial parameters $\theta_0$ may be suboptimal. The problem is to devise a policy that can leverage its own stream of interaction experience at test time, $\{(o_t, a_t, o_{t+1})\}_{t=0}^{N}$, to update its parameters from $\theta_t$ to $\theta_{t+1}$. This update mechanism must be self-supervised, enabling the policy's performance to improve over time in the new environment without any external labels or human intervention.
The core challenge is test-time adaptation to environmental changes and dynamic force shifts. The policy is initially trained offline on a static dataset collected in the source domain. When deployed in a new test environment with different dynamics, the initial parameters $\theta_0$ may be suboptimal. The goal is to devise a policy that leverages its own interaction experience at test time, $\{(o_t, a_t, o_{t+1})\}_{t=0}^{T}$, to update its parameters from $\theta_t$ to $\theta_{t+1}$. This update mechanism must be self-supervised, enabling the policy to improve over time in the new environment without external labels or human intervention.

%-----------------------------------
\subsection{Method Overview}
\label{sec:method_overview}

To address this challenge, we introduce \method (AWP), a framework designed for reactive, self-supervised online adaptation in novel environments. Our key insight is to transform the world model from a passive predictor into an {active supervisor} that drives closed-loop adaptation.

As illustrated on the left of Figure~\ref{fig:overall}, our framework is comprised of three parallel components, all implemented as Diffusion Transformers (DiTs)~\cite{dit_peebles2023scalable} trained through Flow Matching loss~\cite{flowmatching_lipman2022flow}:
\begin{itemize}
    \itemsep0em
    \item A foundational \textbf{World Model}, built upon the powerful, pretrained Cosmos-Predict2~\cite{cosmos2_ali2025world}, which is responsible for predicting future visual states ($x'_{\text{static}}, x'_{\text{gripper}}$).
    \item A lightweight \textbf{Force Predictor}, which complements the visually-focused World Model by extending the system's predictive capabilities into physical dynamics, anticipating future interaction forces ($f'$).
    \item A lightweight \textbf{Action Model}, which serves as the core policy for generating robot actions ($a$).
\end{itemize}
These modules are interconnected via Multi-modal Self-Attention (MMSA)~\cite{sd3_esser2024scaling} to enable deep feature exchange. The entire framework is conditioned on shared inputs—including text embeddings $v_{\text{text}}$ extracted by T5-XXL~\cite{t5_raffel2020exploring}, robot state vector $q$ extracted by an MLP, and the diffusion noise level $\sigma$ (also processed through an MLP)—which are added and injected into each of the three backbones through adaLN~\cite{dit_peebles2023scalable} layers. This unified architecture operates in two distinct modes, determined by the role of the Action Model. In \textbf{Mode I: Action Generation}, it generates an action $a_t$ from Gaussian noise. In \textbf{Mode II: Future Imagination}, it takes an action $a_t$ as input to predict the future state. This dual-mode capability is fundamental to our closed-loop online adaptive learning (AdaOL) strategy, allowing the agent to imagine the consequences of its actions and then correct itself based on real-world outcomes.

In the following sections, we will first detail the World-Model-Driven Diffusion Policy, then describe its extension to force feedback, and finally explain the closed-loop adaptation mechanism.

%-----------------------------------
\subsection{World-Model-Driven Diffusion Policy}
\label{sec:model_and_offline_training}

\paragraph{World Model.}
Our World Model builds upon the pretrained Cosmos-Predict2~\cite{cosmos2_ali2025world}, which we extend to support multi-view video by concatenating the VAE~\cite{cosmos1_agarwal2025cosmos} tokens from each camera view along the temporal dimension. To maintain spatial and temporal coherence across views, we assign each view's token sequence an independent Rotary Position Embedding (RoPE)~\cite{rope_su2024roformer}. Input tokens are paired with a binary mask where `1' indicates a known condition (e.g., $x_{\text{static}}, x_{\text{static}}, f$) and `0' indicates a target for prediction (e.g., a noised version of $x'_{\text{static}}, x'_{\text{gripper}}$).
This mask not only helps the model distinguish between conditional inputs and prediction targets, but is also used after the final denoising step to replace the generated conditional parts with the original inputs, ensuring they are perfectly preserved.

\paragraph{Force Predictor.}
To enhance the agent's capability in contact-rich scenarios, we introduce a dedicated Force Predictor. This module complements the visually-focused World Model by extending the system's predictive power to physical dynamics. It is implemented as a lightweight DiT, structurally similar to the World Model but with significantly fewer parameters (0.4B only). Its role is to predict future force-torque readings ($f'$) based on the current state and action, providing crucial information for tasks involving physical interaction.

\begin{figure}[t]
    \centering
    \includegraphics[width=\linewidth]{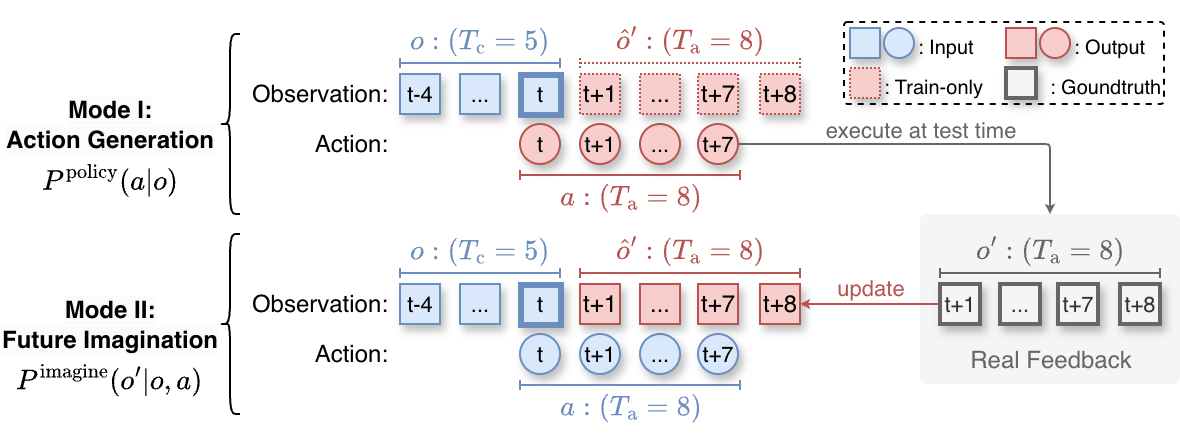}
    \caption{
        % \textbf{Operational modes of our unified model.} Our model can be flexibly configured to perform two distinct tasks using the same underlying architecture.
        \textbf{Input and output details of our unified model in two different modes.}
        \textbf{Mode I (Action Generation):} The model takes an observation history $o$ (e.g., context length $T_c=5$) and predicts a future action sequence $\{a_t, a_{t+1}, \cdots, a_{t+T_\text{a}}\}$ (e.g., action horizon $T_a=8$). At test time, the robot executes this predicted action sequence.
        \textbf{Mode II (Future Imagination):} The model is conditioned on both the observation history $o$ and a ground-truth action sequence $a$, then predicts the corresponding future observation sequence $\hat{o}'$. The discrepancy between this prediction and real environmental feedback is used to update the network parameters during our Adaptive Online Learning (AdaOL) phase.
    }
    \label{fig:stages}
\end{figure}

\paragraph{Action Model and Dual-Mode Operation.}
The Action Model is a lightweight DiT responsible for action generation. Crucially, its input determines the operational mode of the entire AWP framework, a ``switch'' implemented via input masking:
\begin{itemize}
    \itemsep0em
    \item \textbf{Mode I: Action Generation.} To generate an action, the action token input is pure noise, and its corresponding mask is set to all zeros. The model denoises this token using the given observations, effectively computing $a_t$ from $o_t$, which is formulated as:
    \begin{equation}
    \begin{aligned}
        \mathcal{L}_{\text{1}}(\theta) = & \quad \mathbb{E} \big[ \left\| \mathbf{u}_{\theta}(a_k, k, o;\theta) - \mathbf{v}_k(a_k,a) \right\|^2 \big],
    \end{aligned}
    \end{equation}
    where $a_k$ denotes the noised version of $a$ and $v_k$ is the target vector field defined in Flow Matching~\cite{flowmatching_lipman2022flow}.
    \item \textbf{Mode II: Future Imagination.} To predict the future, a known, concrete action $a$ is provided as a condition, and its mask is set to all ones. This conditions the World Model (and Force Predictor) to imagine the future state based on the action, computing $\hat{o}'$ from $o, a$, which can be formulated as:
    \begin{equation}
    \begin{aligned}
        \mathcal{L}_{\text{2}}(\theta) = 
        \mathbb{E} \big[ \left\| \mathbf{u}_{\theta}(o'_k, k, o, a;\theta) - \mathbf{v}_k(o'_k, o') \right\|^2 \big],
    \end{aligned}
    \end{equation}
    where $o'_k$ is a noised version of $o'$.
\end{itemize}
This dual-mode capability is fundamental to our online adaptation strategy, as it allows the agent to first act, then imagine the consequences, and finally compare imagination with reality. Detailed input and output of our unified model in two different modes are shown in Figure~\ref{fig:stages}.

\paragraph{Multi-modal Self-Attention.}
While the three backbones have different functions, they exchange information via MMSA layers. Unlike concatenation, which forces features into a shared space, MMSA allows each module to query information from others while maintaining its own specialized representations. This enables flexible allocation of model capacity (e.g., a large World Model, a small Action Model) and prevents feature corruption. The attention is computed as:
\begin{equation}
\begin{aligned}
    \text{MMSA}(Q, K, V) = A( & [Q_x, Q_f, Q_a], \\
                              & [K_x, K_f, K_a], \\
                              & [V_x, V_f, V_a]),
\end{aligned}
\end{equation}
where A denotes self-attention operation, $[\cdot]$ indicates token-wise concatenation, $Q_x, K_x, V_x$ are the query, key, and value from the World Model, and similarly for the Force Predictor ($f$) and Action Model ($a$).

\paragraph{Joint Training.}
The framework is trained end-to-end with a single Flow Matching loss~\cite{flowmatching_lipman2022flow}. During training, we randomly switch between the two operational modes. With probability $p_{\text{a}}$, we train in \textbf{Action Generation} mode; otherwise, we train in \textbf{Future Imagination} mode. The total loss is the weighted sum of the losses from both modes:
\begin{equation}
\begin{aligned}
    \mathcal{L}_{\text{total}}(\theta) = & \quad p_{\text{a}} \cdot L_1 + (1-p_{\text{a}}) \cdot L_2,
\end{aligned}
\end{equation}
where $\theta$ represents all trainable parameters, $\mathbf{u}_{\theta}$ is the model's predicted vector field, and $\mathbf{v}_k$ is the target vector field at noise step $k$ as defined by Flow Matching. The first term corresponds to the Action Generation loss, where the model learns to predict the vector field for a noised action $a_k$ conditioned on the current observation $o$. The second term is the Future Imagination loss, where the model predicts the vector field for a noised future observation $o'_k$ conditioned on both the current observation $o$ and the action $a$. This joint objective enables the modules to develop a shared understanding of world dynamics, which is the key to enabling our Online Adaptive Learning at test time.

%-----------------------------------
\subsection{Closed-loop Online Adaptive Learning}
\label{sec:adaol}

While offline pre-training provides a strong prior, real-world environments inevitably present visual and physical domain shifts that can degrade performance. To bridge this gap, we introduce Online Adaptive Learning (AdaOL), a closed-loop strategy that enables the agent to reactively self-correct using real-world feedback.

The AdaOL strategy operates as a tight, closed-loop cycle following each action execution, as illustrated in Figure~\ref{fig:teaser}. The process unfolds in the following steps:
\begin{enumerate}
    \itemsep0em 
    \item \textbf{Action Generation:} At timestep $t$, our AWP runs in  action generation mode which takes the current observation $o_t$ as input and generate action $a_t$.
    \item \textbf{Execution:} The robot executes the action $a_t$ in the test environment.
    \item \textbf{Real-world Feedback:} The robot observes the true outcome from the environment, the real future state $o_{t+1}$.
    \item \textbf{Future Imagination:} Concurrently, our AWP runs in future imagination mode which predicts the future state $\hat{o}_{t+1}$ based on the same observation $o_t$ and the executed action $a_t$.
    \item \textbf{Loss and Update:} The core of AdaOL lies in leveraging the discrepancy between reality and prediction. A loss is computed based on the prediction error, typically in a latent space provided by a VAE encoder~\cite{cosmos2_ali2025world}: 
    \begin{equation}
        \mathcal{L}_{\text{AdaOL}} = \| E(o_{t+1}) - E(\hat{o}_{t+1}) \|_2^2,
        \label{eq:adaol_loss}
    \end{equation}
    where $E(\cdot)$ is the encoder. This loss drives an online update by producing a corrective gradient $\Delta w$.
\end{enumerate}
This cycle allows the agent to constantly ground its internal world model (and force predictor) in reality, correcting for any drift or model inaccuracy before the next iteration begins.

To make this online adaptation computationally feasible, we employ Low-Rank Adaptation (LoRA)~\cite{lora_hu2022lora} for the parameter updates. Instead of backpropagating through the entire multi-billion parameter model, the gradients from $\mathcal{L}_{\text{AdaOL}}$ are only used to update only a small set (less than $0.1\%$) of trainable low-rank matrices. This approach significantly reduces the computational and memory overhead at each update step, making the online adaptation practical for deployment on resource-constrained real robots.

\begin{figure}[t]
    \centering
    \includegraphics[width=\linewidth]{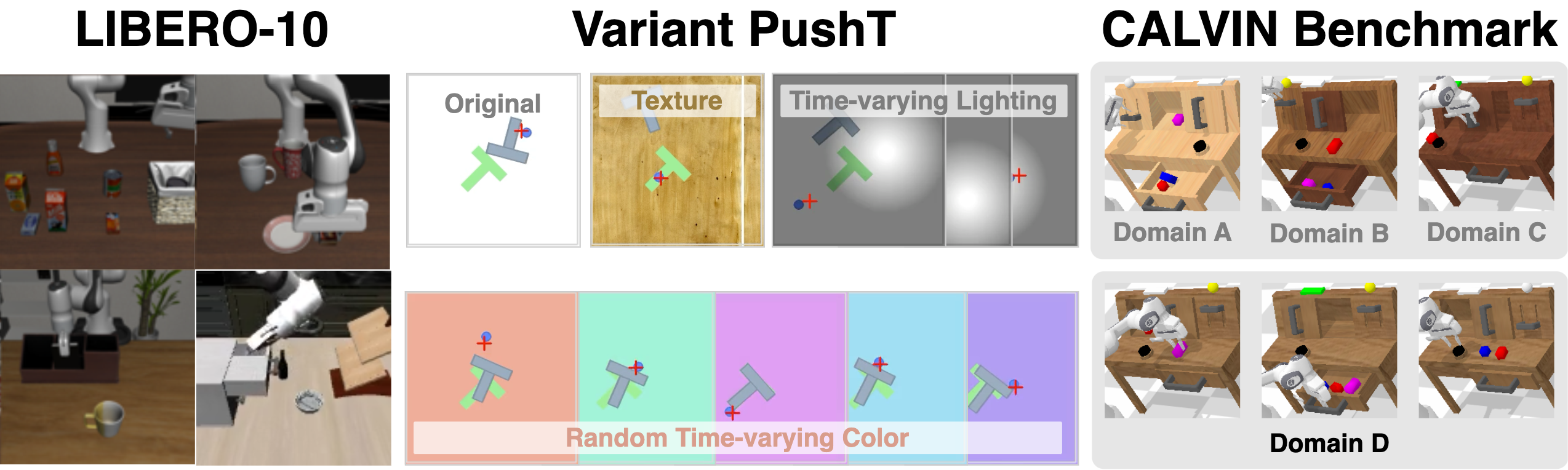}
    \caption{
    Visualizations of the simulated benchmarks used in our experiments: Variant PushT for out-of-distribution robustness, LIBERO for long-horizon skills, and the CALVIN benchmark for language-conditioned tasks across different domains.
    }
    \label{fig:benchmark}
\end{figure}

\section{Experiments}
\label{sec:experiments}

% We conduct extensive experiments to validate the effectiveness of \method across both simulated and real-world manipulation tasks. Our evaluation is designed to answer the following key research questions:
% \begin{itemize}
%     \itemsep0em
%     \item \textbf{Q1:} How does \method compare against state-of-the-art visuomotor policies that do not perform online adaptation?
%     \item \textbf{Q2:} Does our Adaptive Online Learning (AdaOL) strategy effectively improve policy performance under out-of-distribution (OOD) visual and physical shifts?
%     \item \textbf{Q3:} What is the contribution of each component within our proposed framework? (Ablation Study)
% \end{itemize}

%-----------------------------------
\subsection{Experimental Setup}

\paragraph{Simulated Benchmarks.}
As illustrated in Figure~\ref{fig:benchmark}, we evaluate our method on three diverse simulated benchmarks. 
\textbf{(1) LIBERO-10}~\cite{libero_liu2024libero}: A benchmark for long-horizon, compositional skills across various household scenarios. It tests the policy's ability to handle complex task sequences.
\textbf{(2) Variant PushT}~\cite{diffusionplanning_janner2022planning}: We modify the classic PushT task to specifically test out-of-distribution (OOD) robustness. After training on the original setting, we evaluate the policy's ability to adapt to test-time variations in background texture, random time-varying lighting, and random time-varying color. Task success is measured by the Intersection over Union (IoU) between the goal region and the final object position.
\textbf{(3) CALVIN}~\cite{calvin_mees2022calvin}: The CALVIN benchmark consists of four environments, A, B, C, and D, each belonging to different domains. 
Each domain provides 6 hours of human teleoperated data across 34 different tasks. 
To validate the cross-domain adaptive ability of methods, we only use the ABC$\rightarrow$D evaluation protocol.
In each sequence, the robot is required to continuously solve five tasks in a row.

\begin{figure*}[ht]
    \centering
    \includegraphics[width=\textwidth]{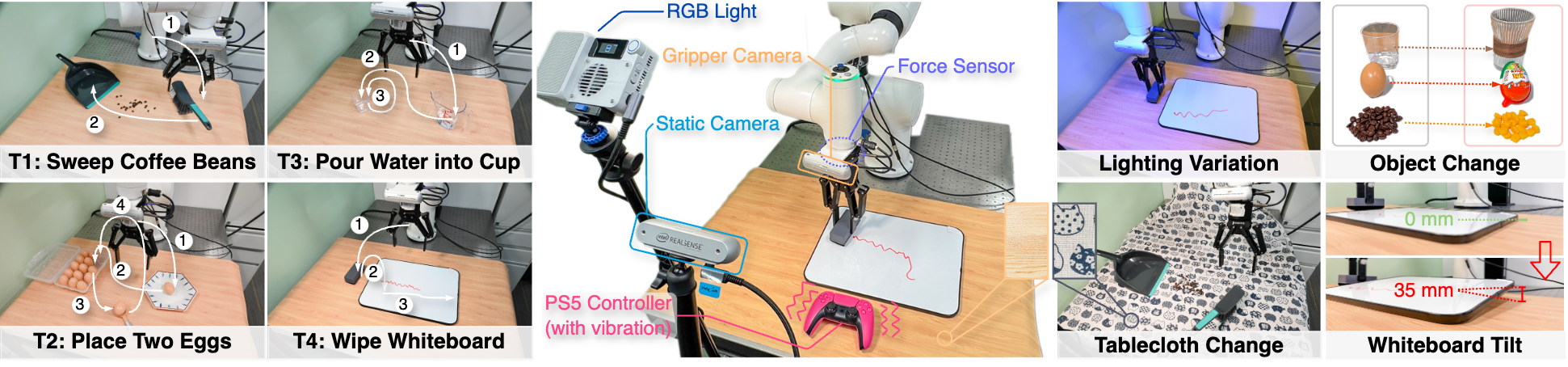}
    \caption{
    \textbf{Overview of our real-robot evaluation.} Our experimental setup (center) features an INOVO robotic arm with multi-modal sensing capabilities, including gripper/static cameras and a force sensor. We test our \method on four diverse manipulation tasks (left), such as sweeping beans and placing eggs. To specifically evaluate the effectiveness of our AdaOL strategy, we introduce a variety of challenging out-of-distribution (OOD) shifts during execution (right). These include visual perturbations like drastic lighting and texture changes, as well as physical perturbations like swapping objects and altering the workspace geometry (e.g., tilting the whiteboard).
    }
    \label{fig:real}
\end{figure*}

\paragraph{Real-World Setup.}
Our real-world experiments are conducted using a 6-DoF robot arm equipped with a gripper camera and a wrist-mounted force-torque sensor, supplemented by a static third-person camera (Figure~\ref{fig:real}). For offline training, we collected a dataset of human demonstrations via teleoperation with a PS5 controller. The controller provided haptic feedback by mapping force sensor readings to vibrations, allowing the operator to feel the interaction forces.

We evaluate our method on four challenging, long-horizon tasks requiring both precise motion and rich physical interaction: (1) sweeping coffee beans into a dustpan, (2) picking and placing two eggs into a box, (3) pouring water from a measuring cup, and (4) wiping a whiteboard. To specifically test our online adaptive learning strategy, we introduce various domain shifts during deployment. These include visual perturbations (e.g., changing lighting, tablecloth textures) and physical perturbations (e.g., substituting objects with different weights, altering the whiteboard's incline). Further experimental details are available in the appendix.

% \paragraph{Baselines.}
% We compare \method against several strong baselines. For simulated benchmarks, we compare with \textbf{Octo}~\cite{shah2023octo}, a state-of-the-art generalist policy, and a \textbf{No-Adaptation} version of our own model which is pre-trained but does not perform online updates with AdaOL. For real-world experiments, we compare against a fine-tuned \textbf{RT-2}~\cite{brohan2023rt2} model as a representative of large-scale VLA models.

\begin{table}[t]
\centering
\caption{Comparison of success rates on the LIBERO-10 benchmark, under two settings: with only a static camera, and with full multi-modal inputs. Note: AWP is the short name of our method AdaWorldPolicy without online adaptive learning.}
\label{tab:libero_results}
\resizebox{\columnwidth}{!}{%
\begin{tabular}{lccccc}
\toprule
\textbf{Methods} & \textbf{Pub.} & \textbf{Static Camera} & \textbf{Gripper Camera} & \textbf{Joint States} & \textbf{Success} \\
\midrule
UVA~\cite{uva_li2025unified} & RSS'25        & \checkmark &            &            & 0.89 \\
\textbf{AWP} & \textbf{Ours} & \checkmark &            &            & \textbf{0.91} \\
\midrule
OpenVLA~\cite{openvla_kim2024openvla} & CoRL'24    & \checkmark & \checkmark & \checkmark & 0.54 \\
SpatialVLA~\cite{spatialvla_qu2025spatialvla} & RSS'25 & \checkmark & \checkmark & \checkmark & 0.56 \\
Pi0-fast~\cite{pi0fast_pertsch2025fast} & RSS'25   & \checkmark & \checkmark & \checkmark & 0.60 \\
FlowVLA~\cite{flowvla_zhong2025flowvla} & arXiv'25  & \checkmark & \checkmark & \checkmark & 0.73 \\
MODE~\cite{mode_reuss2024efficient} & NIPS'24      & \checkmark & \checkmark & \checkmark & 0.94 \\
OpenVLA-OFT~\cite{openvla-oft_kim2025fine} & RSS'25      & \checkmark & \checkmark & \checkmark & 0.94 \\
\textbf{AWP} & \textbf{Ours} & \checkmark & \checkmark & \checkmark & \textbf{0.96} \\
\bottomrule
\end{tabular}%
}
\vspace{-0.3em}
\end{table}
%-----------------------------------
\subsection{Implementation Details}
We implement our model in PyTorch, building upon the publicly available Cosmos-Predict2 architecture~\cite{cosmos2_ali2025world}. Our full \method model consists of a 2B parameter world model and two 0.4B parameter DiTs for action and force prediction. For simulator benchmarks without force data, we remove the force predictor.

\textbf{Offline training.} For the offline training stage, we largely follow the training recipe of Cosmos-Predict2. We train the model using the AdamW optimizer~\cite{adamw_loshchilov2017decoupled} on 8 A100 (80GB) GPUs. To maximize hardware utilization, we use the largest possible per-GPU batch size, resulting in a global batch size that varies between 64 and 256 across different datasets. The initial learning rate is set to $1 \times 10^{-4}$. Once the training loss plateaus, we apply a linear decay schedule for a maximum of 20k steps, reducing the learning rate to 1\% of its initial value. 
% The model is trained to predict a sequence of future actions corresponding to a horizon of approximately 1 second, with the exact number of steps varying based on the dataset's frame rate.

\textbf{Online Learning.} We perform test-time adaptation on a single NVIDIA RTX 5880 (48GB) GPU. To keep the update process lightweight, we employ LoRA with rank 16, applying adaptable matrices only to the first 4 layers of each backbone. For each incoming data sample (effective batch size of 1), we perform two gradient descent steps with a small, constant learning rate of $5 \times 10^{-7}$. This targeted, lightweight update strategy minimizes computational overhead. As a result, the average inference speed with TTA enabled is only approximately 5\% slower than the baseline without adaptation.

Further details on the network architecture, data processing pipelines for each benchmark, and a complete list of hyperparameters can be found in the Appendix.

%-----------------------------------
\subsection{Main Results}

\paragraph{LIBERO-10 Results.}
As shown in Table~\ref{tab:libero_results}, our base {AWP} architecture achieves a new state-of-the-art results on the LIBERO-10 benchmark. It outperforms prior methods like UVA and MODE in both unimodal (static camera) and full multi-modal settings, demonstrating the strength of our MMSA-based design even without online adaptation.

\paragraph{Variant PushT Results.}
The Variant PushT benchmark (Table~\ref{tab:pusht_results}) highlights the critical role of our online adaptation. While the offline-trained {AWP} degrades significantly under visual domain shifts, activating online learning ({AWP (ol)}) consistently and substantially recovers performance. This makes AWP (ol) the top-performing method across all challenging out-of-distribution scenarios, demonstrating the effectiveness of our TTA mechanism.

\paragraph{CALVIN Results.}
On the long-horizon CALVIN benchmark (Table~\ref{tab:calvin_results}), our base {AWP} model again achieves state-of-the-art results, outperforming all prior methods in long-sequence task completion. Enabling online learning provides a further consistent performance boost, demonstrating that our TTA strategy can effectively fine-tune an already well-generalized policy.

\begin{table}[t]
\centering
\caption{Performance on the Variant PushT benchmark under different distribution shifts. We report the mean Intersection-over-Union (IoU) results. Note: AWP/AWP (ol) is the short name of our method AdaWorldPolicy without/with adaptive online learning.}
\label{tab:pusht_results}
\resizebox{\columnwidth}{!}{%
\begin{tabular}{lccccc}
\toprule
\textbf{Methods} & \textbf{Pub.} & \textbf{Original} & \textbf{Texture} & \textbf{Rand Light} & \textbf{Rand Color} \\
\midrule
DP~\cite{diffusionpolicy_chi2023diffusion}  &   IJRR'24           & 0.78          & 0.18          & 0.14 & 0.11 \\
OpenVLA~\cite{openvla_kim2024openvla}   & CoRL'24      & 0.35          & 0.22          & 0.20 & 0.14 \\
UniPi~\cite{unipi_du2023learning}   & NIPS'23        & 0.42          & 0.35          & 0.33 & 0.18 \\
UVA~\cite{uva_li2025unified}     & RSS'25        & 0.94          & 0.11          & 0.54 & 0.13 \\
\midrule
\textbf{AWP} & {Ours} & 0.97          & 0.47          & 0.71 & 0.61 \\
\textbf{AWP (ol)} & {Ours} & \textbf{0.98} & \textbf{0.51} & \textbf{0.77} & \textbf{0.66} \\
\bottomrule
\end{tabular}%
}
\end{table}

\begin{table}[t]
\centering
\caption{Performance on the CALVIN benchmark for the cross-domain setting (ABC for training, D for evaluation). We report the success rate (\%) for different task lengths (the number of consecutive instructions) and the average length of successfully completed sub-sequences (Avg. Len.). We compare with the baseline methods without pretraining on any extra human demonstration data outside the benchmark. Note: AWP/AWP (ol) is the short name of our method AdaWorldPolicy without/with adaptive online learning.}
\label{tab:calvin_results}
\resizebox{\linewidth}{!}{%
\begin{tabular}{lcccccccc}
\toprule
\multirow{2}{*}{\textbf{Methods}} & \multirow{2}{*}{\textbf{Pub.}} & \multicolumn{5}{c}{\textbf{Success Rate (\%) for Task Length}} & \multirow{2}{*}{\textbf{Avg. Len.}} \\
\cmidrule(lr){3-7}
& & \textbf{1} & \textbf{2} & \textbf{3} & \textbf{4} & \textbf{5} & \\
\midrule
DP~\cite{diffusionpolicy_chi2023diffusion}            & IJRR'24    & 62.2 & 30.9 & 13.2 & 5.0  & 1.6  & 1.13$\pm$0.02 \\
MDT~\cite{mdt_reuss2024multimodal}             & RSS'24    & 61.7 & 40.6 & 23.8 & 14.7 & 8.7  & 1.54$\pm$0.04 \\
RoboFlamingo~\cite{roboflamingo_bytedance2024iclr}    & ICLR'24   & 82.4 & 61.9 & 46.6 & 33.1 & 23.5 & 2.47$\pm$0.00 \\
GR-1~\cite{gr1_bytedance2024iclr}            & ICLR'24   & 85.4 & 71.2 & 59.6 & 49.7 & 40.1 & 3.06$\pm$0.00 \\
OpenVLA~\cite{openvla_kim2024openvla}         & CoRL'24   & 91.3 & 77.8 & 62.0 & 52.1 & 43.5 & 3.27$\pm$0.00 \\
MoDE~\cite{mode_reuss2024efficient}            & ICLR'25   & 91.5 & 79.2 & 67.3 & 55.8 & 45.3 & 3.39$\pm$0.03 \\
GR-MG~\cite{grmg_bytedance2024}           & RAL'25    & 91.0 & 79.1 & 67.8 & 56.9 & 47.7 & 3.42$\pm$0.28 \\
\midrule
\textbf{AWP} & {Ours} & 91.8 & 79.2 & 68.5 & 62.8 & 48.0 & 3.51$\pm$0.03 \\
\textbf{AWP (ol)} & {Ours} & \textbf{92.0} & \textbf{79.6} & \textbf{68.6} & \textbf{63.0} & \textbf{48.0} & \textbf{3.54$\pm$0.04} \\
\bottomrule
\end{tabular}%
}
\end{table}

\begin{figure*}[t]
    \centering
    \includegraphics[width=0.9\textwidth]{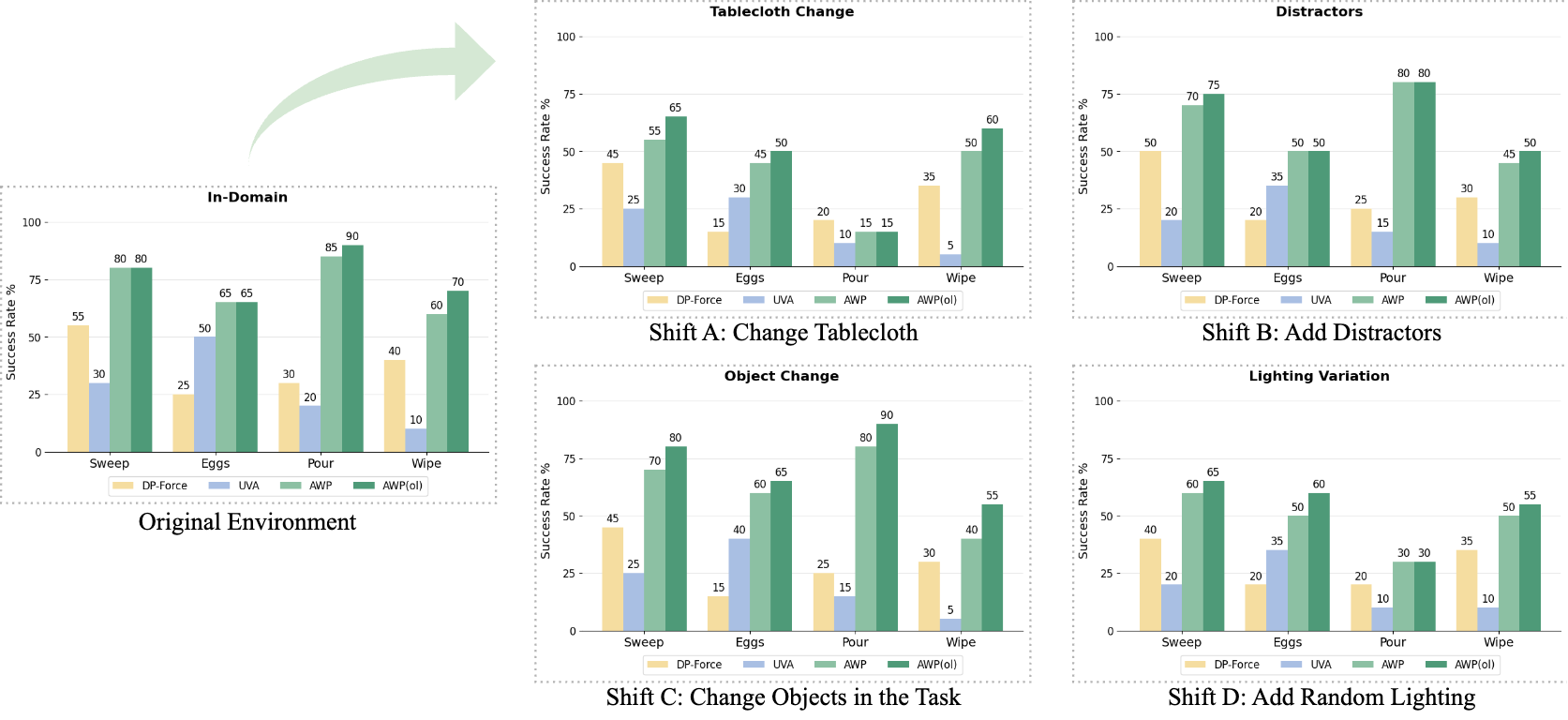}
    \caption{\textbf{Real-world evaluation results under domain shifts.} Our base model, AdaWorldPolicy (or AWP in short), shows a strong performance in the original environment (left). When suffering from various visual and physical shifts at test time (right), our full model with online adaptive learning, \textbf{AWP (ol)}, consistently and significantly improves the success rate, showcasing robust online adaptation.}
    \label{fig:real_exp_curve}
\end{figure*}

\paragraph{Performance in Real-World Experiments.}
We validate our approach on a physical robot across four challenging long-horizon tasks, testing it to various out-of-distribution (OOD) shifts at test time. The results are summarized in Figure~\ref{fig:real_exp_curve}. In the original in-domain environment (left), our offline-trained {AWP} model already establishes a strong performance, outperforming the DP-Force and UVA baselines on most tasks.

The critical advantage of our method emerges under the four OOD scenarios (right). While the performance of all methods degrades under these shifts, our full model with online learning, {AWP (ol)}, consistently and significantly outperforms its offline-only counterpart ({AWP}) and all other baselines. For instance, under the ``Object Change" shift, our AWP (ol) improves the success rate on the ``Pour" task from 80\% to 90\%. This consistent performance gain across all tasks and perturbation types validates the effectiveness of our Test-Time Adaptation strategy. The entire closed-loop process, including action generation, online updating, and device latency, runs at an average of 4Hz.

\begin{table}[t]
\centering
\caption{Ablation study on the main components of \method{}. We report the average success rate (\%) across four real-world in-domain tasks. Our full method with online adaptive learning achieves the best performance, while removing or replacing any key component leads to a significant performance drop, which validates our design choices.}
\label{tab:ablation}
\resizebox{0.72\linewidth}{!}{%
\begin{tabular}{lc}
\toprule
\textbf{Configuration} & \textbf{Success Rate (\%)} \\
\midrule
\method{} w/ AdaOL (Full) & \textbf{76.3} \\
\method{} w/o AdaOL & 72.5 \\
\midrule
w/o Force Predictor & 53.8 \\
w/o World Model Supervision & 46.3 \\
\midrule
MMSA $\rightarrow$ Concatenation & 36.3 \\
MMSA $\rightarrow$ Cross-Attention & 50.0 \\
\bottomrule
\end{tabular}%
}
\vspace{-0.1em}
\end{table}
%-----------------------------------
\subsection{Ablation Studies}
\label{sec:ablation}

We conduct ablation studies to validate our design choices, with results reported as average success rates across four real-world in-domain tasks in Table~\ref{tab:ablation}. Our full model, \method with AdaOL, achieves the highest score of 76.3\%. Disabling test-time online learning (\method w/o AdaOL) reduces the success rate to 72.5\%, confirming the value of continuous adaptation. Removing the Force Predictor module causes a significant drop to 53.8\%, highlighting the necessity of modeling physical contact dynamics. The most critical component is the world model's supervision without its training loss, the framework degenerates into a behavioral cloning policy, and performance plummets to 46.3\%. This validates our core premise of using the world model as an active supervisor. Finally, replacing our Multi-modal Self-Attention (MMSA) layers with simpler fusion methods, such as concatenation (36.3\%) or cross-attention (50.0\%), severely degrades performance, confirming that MMSA is superior for integrating the different modules while preserving their specialized representations.
\section{Conclusion}
\label{sec:conclusion}

We have introduced \method, a unified framework for robotic manipulation that integrates a world-model-driven diffusion policy with a novel online adaptive learning strategy (AdaOL). This strategy uses prediction errors from the improved world model as a self-supervised signal, driving efficient LoRA-based updates to continuously reduce visual and physical domain shifts. Extensive experiments show that our \method with online adaptive learning significantly outperforms strong non-adaptive baselines under out-of-distribution conditions in both simulation and real-world settings, achieving robust performance with minimal computational latency. Future work will extend this adaptation mechanism to further address long-horizon planning failures and scale it to larger network architectures.
{
    \small
    \bibliographystyle{ieeenat_fullname}
    \bibliography{main}

@String(CVPR= {IEEE Conf. Comput. Vis. Pattern Recog.})

@String(ICCV= {Int. Conf. Comput. Vis.})

@String(NIPS= {Adv. Neural Inform. Process. Syst.})

@String(ICLR = {Int. Conf. Learn. Represent.})

@String(CVPR  = {CVPR})

@String(ICCV  = {ICCV})

@String(NIPS  = {NeurIPS})

@String(ICLR  = {ICLR})

@inproceedings{gr1_bytedance2024iclr,
  title={Unleashing Large-Scale Video Generative Pre-training for Visual Robot Manipulation},
  author={Wu, Hongtao and Jing, Ya and Cheang, Chilam and Chen, Guangzeng and Xu, Jiafeng and Li, Xinghang and Liu, Minghuan and Li, Hang and Kong, Tao},
  booktitle={ICLR},
  year={2024},
}

@inproceedings{roboflamingo_bytedance2024iclr,
  title={{Vision-Language} Foundation Models as Effective Robot Imitators},
  author={Li, Xinghang and Liu, Minghuan and Zhang, Hanbo and Yu, Cunjun and Xu, Jie and Wu, Hongtao and Cheang, Chilam and Jing, Ya and Zhang, Weinan and Liu, Huaping and others},
  booktitle={ICLR},
  year={2024},
}

@article{grmg_bytedance2024,
  title={{GR-MG}: Leveraging Partially Annotated Data via Multi-Modal Goal Conditioned Policy},
  author={Li, Peiyan and Wu, Hongtao and Huang, Yan and Cheang, Chilam and Wang, Liang and Kong, Tao},
  journal={arXiv preprint arXiv:2408.14368},
  year={2024}
}

@article{eai_llm_tan2024true,
  title={True knowledge comes from practice: Aligning llms with embodied environments via reinforcement learning},
  author={Tan, Weihao and Zhang, Wentao and Liu, Shanqi and Zheng, Longtao and Wang, Xinrun and An, Bo},
  journal={arXiv preprint arXiv:2401.14151},
  year={2024}
}

@inproceedings{eai_llm_szot2023large,
  title={Large language models as generalizable policies for embodied tasks},
  author={Szot, Andrew and Schwarzer, Max and Agrawal, Harsh and Mazoure, Bogdan and Metcalf, Rin and Talbott, Walter and Mackraz, Natalie and Hjelm, R Devon and Toshev, Alexander T},
  booktitle={ICLR},
  year={2023}
}

@article{eai_llm_zheng2023steve,
  title={{Steve-Eye}: Equipping llm-based embodied agents with visual perception in open worlds},
  author={Zheng, Sipeng and Liu, Jiazheng and Feng, Yicheng and Lu, Zongqing},
  journal={arXiv preprint arXiv:2310.13255},
  year={2023}
}

@article{rt1_brohan2022rt,
  title={{RT-1}: Robotics transformer for real-world control at scale},
  author={Brohan, Anthony and Brown, Noah and Carbajal, Justice and Chebotar, Yevgen and Dabis, Joseph and Finn, Chelsea and Gopalakrishnan, Keerthana and Hausman, Karol and Herzog, Alex and Hsu, Jasmine and others},
  journal={arXiv preprint arXiv:2212.06817},
  year={2022}
}

@article{rt2_brohan2023rt,
  title={{RT-2}: Vision-language-action models transfer web knowledge to robotic control},
  author={Brohan, Anthony and Brown, Noah and Carbajal, Justice and Chebotar, Yevgen and Chen, Xi and Choromanski, Krzysztof and Ding, Tianli and Driess, Danny and Dubey, Avinava and Finn, Chelsea and others},
  journal={arXiv preprint arXiv:2307.15818},
  year={2023}
}

@article{libero_liu2024libero,
  title={{LIBERO}: Benchmarking knowledge transfer for lifelong robot learning},
  author={Liu, Bo and Zhu, Yifeng and Gao, Chongkai and Feng, Yihao and Liu, Qiang and Zhu, Yuke and Stone, Peter},
  journal={NeurIPS},
  volume={36},
  year={2024}
}

@article{calvin_mees2022calvin,
  title={{CALVIN}: A benchmark for language-conditioned policy learning for long-horizon robot manipulation tasks},
  author={Mees, Oier and Hermann, Lukas and Rosete-Beas, Erick and Burgard, Wolfram},
  journal={IEEE Robotics and Automation Letters},
  volume={7},
  number={3},
  pages={7327--7334},
  year={2022},
  publisher={IEEE}
}

@article{diffusionplanning_janner2022planning,
  title={Planning with diffusion for flexible behavior synthesis},
  author={Janner, Michael and Du, Yilun and Tenenbaum, Joshua B and Levine, Sergey},
  journal={arXiv preprint arXiv:2205.09991},
  year={2022}
}

@article{diffusionpolicy_chi2023diffusion,
  title={{Diffusion Policy}: Visuomotor policy learning via action diffusion},
  author={Chi, Cheng and Xu, Zhenjia and Feng, Siyuan and Cousineau, Eric and Du, Yilun and Burchfiel, Benjamin and Tedrake, Russ and Song, Shuran},
  journal={The International Journal of Robotics Research},
  pages={02783649241273668},
  year={2024},
  publisher={SAGE Publications Sage UK: London, England}
}

@inproceedings{mdt_reuss2024multimodal,
  title={Multimodal Diffusion Transformer: Learning Versatile Behavior from Multimodal Goals},
  author={Reuss, Moritz and Yagmurlu, Omer Erdinc and Wenzel, Fabian and Lioutikov, Rudolf},
  booktitle={Robotics: Science and Systems},
  year={2024}
}

@inproceedings{adaln_huang2017arbitrary,
  title={Arbitrary style transfer in real-time with adaptive instance normalization},
  author={Huang, Xun and Belongie, Serge},
  booktitle={ICCV},
  pages={1501--1510},
  year={2017}
}

@inproceedings{mode_reuss2024efficient,
  title={Efficient diffusion transformer policies with mixture of expert denoisers for multitask learning},
  author={Reuss, Moritz and Pari, Jyothish and Agrawal, Pulkit and Lioutikov, Rudolf},
  booktitle={ICLR},
  year={2025}
}

@inproceedings{spatialvla_qu2025spatialvla,
  title={Spatialvla: Exploring spatial representations for visual-language-action model},
  author={Qu, Delin and Song, Haoming and Chen, Qizhi and Yao, Yuanqi and Ye, Xinyi and Ding, Yan and Wang, Zhigang and Gu, JiaYuan and Zhao, Bin and Wang, Dong and others},
  booktitle={Robotics: Science and Systems},
  year={2025}
}

@inproceedings{openvla-oft_kim2025fine,
  title={Fine-tuning vision-language-action models: Optimizing speed and success},
  author={Kim, Moo Jin and Finn, Chelsea and Liang, Percy},
  booktitle={Robotics: Science and Systems},
  year={2025}
}

@inproceedings{octo_team2024octo,
  title={Octo: An open-source generalist robot policy},
  author={Team, Octo Model and Ghosh, Dibya and Walke, Homer and Pertsch, Karl and Black, Kevin and Mees, Oier and Dasari, Sudeep and Hejna, Joey and Kreiman, Tobias and Xu, Charles and others},
  booktitle={Robotics: Science and Systems},
  year={2024}
}

@inproceedings{pi0_black2410pi0,
  title={$\pi$0: A vision-language-action flow model for general robot control. CoRR, abs/2410.24164, 2024. doi: 10.48550},
  author={Black, Kevin and Brown, Noah and Driess, Danny and Esmail, Adnan and Equi, Michael and Finn, Chelsea and Fusai, Niccolo and Groom, Lachy and Hausman, Karol and Ichter, Brian and others},
  booktitle={Robotics: Science and Systems},
  year={2025}
}

@inproceedings{pi0fast_pertsch2025fast,
  title={Fast: Efficient action tokenization for vision-language-action models},
  author={Pertsch, Karl and Stachowicz, Kyle and Ichter, Brian and Driess, Danny and Nair, Suraj and Vuong, Quan and Mees, Oier and Finn, Chelsea and Levine, Sergey},
  booktitle={Robotics: Science and Systems},
  year={2025},
}

@article{flowvla_zhong2025flowvla,
  title={Flowvla: Thinking in motion with a visual chain of thought},
  author={Zhong, Zhide and Yan, Haodong and Li, Junfeng and Liu, Xiangchen and Gong, Xin and Song, Wenxuan and Chen, Jiayi and Li, Haoang},
  journal={arXiv e-prints},
  pages={arXiv--2508},
  year={2025}
}

@InProceedings{openvla_kim2024openvla,
  title = 	 {OpenVLA: An Open-Source Vision-Language-Action Model},
  author =       {Kim, Moo Jin and Pertsch, Karl and Karamcheti, Siddharth and Xiao, Ted and Balakrishna, Ashwin and Nair, Suraj and Rafailov, Rafael and Foster, Ethan P and Sanketi, Pannag R and Vuong, Quan and Kollar, Thomas and Burchfiel, Benjamin and Tedrake, Russ and Sadigh, Dorsa and Levine, Sergey and Liang, Percy and Finn, Chelsea},
  booktitle = 	 {Proceedings of The 8th Conference on Robot Learning},
  pages = 	 {2679--2713},
  year = 	 {2025},
}

@inproceedings{uva_li2025unified,
  title={Unified video action model},
  author={Li, Shuang and Gao, Yihuai and Sadigh, Dorsa and Song, Shuran},
  booktitle={Robotics: Science and Systems},
  year={2025}
}

@inproceedings{unipi_du2023learning,
  title={Learning Universal Policies via Text-Guided Video Generation},
  author={Du, Yilun and Yang, Mengjiao and Dai, Bo and Dai, Hanjun and Nachum, Ofir and Tenenbaum, Joshua B and Schuurmans, Dale and Abbeel, Pieter},
  booktitle={NIPS},
  pages={arXiv--2302},
  year={2023}
}

@article{cosmos1_agarwal2025cosmos,
  title={Cosmos world foundation model platform for physical ai},
  author={Agarwal, Niket and Ali, Arslan and Bala, Maciej and Balaji, Yogesh and Barker, Erik and Cai, Tiffany and Chattopadhyay, Prithvijit and Chen, Yongxin and Cui, Yin and Ding, Yifan and others},
  journal={arXiv preprint arXiv:2501.03575},
  year={2025}
}

@article{cosmos2_ali2025world,
  title={World Simulation with Video Foundation Models for Physical AI},
  author={Ali, Arslan and Bai, Junjie and Bala, Maciej and Balaji, Yogesh and Blakeman, Aaron and Cai, Tiffany and Cao, Jiaxin and Cao, Tianshi and Cha, Elizabeth and Chao, Yu-Wei and others},
  journal={arXiv preprint arXiv:2511.00062},
  year={2025}
}

@inproceedings{zhu2025unified,
  title={Unified world models: Coupling video and action diffusion for pretraining on large robotic datasets},
  author={Zhu, Chuning and Yu, Raymond and Feng, Siyuan and Burchfiel, Benjamin and Shah, Paarth and Gupta, Abhishek},
  booktitle={Robotics: Science and Systems},
  year={2025}
}

@inproceedings{sd3_esser2024scaling,
  title={Scaling rectified flow transformers for high-resolution image synthesis},
  author={Esser, Patrick and Kulal, Sumith and Blattmann, Andreas and Entezari, Rahim and M{\"u}ller, Jonas and Saini, Harry and Levi, Yam and Lorenz, Dominik and Sauer, Axel and Boesel, Frederic and others},
  booktitle={ICML},
  year={2024}
}

@article{worldvla_cen2025worldvla,
  title={WorldVLA: Towards Autoregressive Action World Model},
  author={Cen, Jun and Yu, Chaohui and Yuan, Hangjie and Jiang, Yuming and Huang, Siteng and Guo, Jiayan and Li, Xin and Song, Yibing and Luo, Hao and Wang, Fan and others},
  journal={arXiv preprint arXiv:2506.21539},
  year={2025}
}

@inproceedings{enerverseac_jiang2025enerverse,
  title={Enerverse-ac: Envisioning embodied environments with action condition},
  author={Jiang, Yuxin and Chen, Shengcong and Huang, Siyuan and Chen, Liliang and Zhou, Pengfei and Liao, Yue and He, Xindong and Liu, Chiming and Li, Hongsheng and Yao, Maoqing and others},
  booktitle={Robotics: Science and Systems},
  year={2025}
}

@inproceedings{naviworldmodel_bar2025navigationWM,
  title        = {Navigation World Models},
  author       = {Bar, Amir and Zhou, Gaoyue and Tran, Danny and Darrell, Trevor and LeCun, Yann},
  booktitle    = {CVPR},
  year         = {2025},
}

@article{dinowm_zhou2024dino,
  title={Dino-wm: World models on pre-trained visual features enable zero-shot planning},
  author={Zhou, Gaoyue and Pan, Hengkai and LeCun, Yann and Pinto, Lerrel},
  journal={arXiv preprint arXiv:2411.04983},
  year={2024}
}

@article{jepa2_assran2025vjepa2,
  title        = {V‑JEPA 2: Self‑Supervised Video Models Enable Understanding, Prediction and Planning},
  author       = {Assran, Mido and Bardes, Adrien and Fan, David and Garrido, Quentin and Howes, Russell and Komeili, Mojtaba and Muckley, Matthew and Rizvi, Ammar and Roberts, Claire and Sinha, Koustuv and Zholus, Artem and Arnaud, Sergio and Gejji, Abha and Martin, Ada and Hogan, Francois Robert and Dugas, Daniel and Bojanowski, Piotr and Khalidov, Vasil and Labatut, Patrick and Massa, Francisco and Szafraniec, Marc and Krishnakumar, Kapil and Li, Yong and Ma, Xiaodong and Chandar, Sarath and Meier, Franziska and LeCun, Yann and Rabbat, Michael and Ballas, Nicolas},
  journal      = {arXiv preprint arXiv:2506.09985},
  year         = {2025},
}

@inproceedings{flowmatching_lipman2022flow,
  title={Flow matching for generative modeling},
  author={Lipman, Yaron and Chen, Ricky TQ and Ben-Hamu, Heli and Nickel, Maximilian and Le, Matt},
  booktitle={ICLR},
  year={2023}
}

@inproceedings{dit_peebles2023scalable,
  title={Scalable diffusion models with transformers},
  author={Peebles, William and Xie, Saining},
  booktitle={CVPR},
  pages={4195--4205},
  year={2023}
}

@article{t5_raffel2020exploring,
  title={Exploring the limits of transfer learning with a unified text-to-text transformer},
  author={Raffel, Colin and Shazeer, Noam and Roberts, Adam and Lee, Katherine and Narang, Sharan and Matena, Michael and Zhou, Yanqi and Li, Wei and Liu, Peter J},
  journal={Journal of machine learning research},
  volume={21},
  number={140},
  pages={1--67},
  year={2020}
}

@article{rope_su2024roformer,
  title={Roformer: Enhanced transformer with rotary position embedding},
  author={Su, Jianlin and Ahmed, Murtadha and Lu, Yu and Pan, Shengfeng and Bo, Wen and Liu, Yunfeng},
  journal={Neurocomputing},
  volume={568},
  pages={127063},
  year={2024},
  publisher={Elsevier}
}

@article{adamw_loshchilov2017decoupled,
  title={Decoupled weight decay regularization},
  author={Loshchilov, Ilya and Hutter, Frank},
  journal={arXiv preprint arXiv:1711.05101},
  year={2017}
}

@inproceedings{TST_sun2020test,
  title={Test-time training with self-supervision for generalization under distribution shifts},
  author={Sun, Baochen and Wei, Yang and Feng, Chen and Wang, Bo and D'Souza, Sophia and Hoiem, Derek},
  booktitle={Proceedings of the 37th International Conference on Machine Learning (ICML)},
  year={2020},
  pages={9336--9346},
}

@inproceedings{lora_hu2022lora,
  title={Lora: Low-rank adaptation of large language models.},
  author={Hu, Edward J and Shen, Yelong and Wallis, Phillip and Allen-Zhu, Zeyuan and Li, Yuanzhi and Wang, Shean and Wang, Lu and Chen, Weizhu and others},
  booktitle={ICLR},
  year={2022}
}

@inproceedings{hafner2020dreamer,
  title={Dream to Control: Learning Behaviors by Latent Imagination},
  author={Hafner, Danijar and Lillicrap, Timothy and Ba, Jimmy and Norouzi, Mohammad},
  booktitle={ICLR},
  year={2020}
}

@article{hafner2025dreamerv3,
  title={Mastering diverse control tasks through world models},
  author={Hafner, Danijar and Pasukonis, Jurgis and Ba, Jimmy and Lillicrap, Timothy},
  journal={Nature},
  pages={1--7},
  year={2025},
  publisher={Nature Publishing Group}
}

@article{schrittwieser2020mastering,
  title={Mastering atari, go, chess and shogi by planning with a learned model},
  author={Schrittwieser, Julian and Antonoglou, Ioannis and Hubert, Thomas and Simonyan, Karen and Sifre, Laurent and Schmitt, Simon and Guez, Arthur and Lockhart, Edward and Hassabis, Demis and Graepel, Thore and others},
  journal={Nature},
  volume={588},
  number={7839},
  pages={604--609},
  year={2020},
  publisher={Nature Publishing Group UK London}
}

@inproceedings{ajay2022conditional,
  title={Is conditional generative modeling all you need for decision-making?},
  author={Ajay, Anurag and Du, Yilun and Gupta, Abhi and Tenenbaum, Joshua and Jaakkola, Tommi and Agrawal, Pulkit},
  booktitle={ICLR},
  year={2023}
}

@inproceedings{wang2022continual,
  title={Continual test-time domain adaptation},
  author={Wang, Qin and Fink, Olga and Van Gool, Luc and Dai, Dengxin},
  booktitle={Proceedings of the IEEE/CVF Conference on Computer Vision and Pattern Recognition},
  pages={7201--7211},
  year={2022}
}

@article{adaptdiffuser_liang2023adaptdiffuser,
  title={Adaptdiffuser: Diffusion models as adaptive self-evolving planners},
  author={Liang, Zhixuan and Mu, Yao and Ding, Mingyu and Ni, Fei and Tomizuka, Masayoshi and Luo, Ping},
  journal={arXiv preprint arXiv:2302.01877},
  year={2023}
}

@inproceedings{ma2024hierarchicalDiffusionPolicy,
  title        = {Hierarchical Diffusion Policy for Kinematics‑Aware Multi‑Task Robotic Manipulation},
  author       = {Ma, Yueen and Song, Zixing and Zhuang, Yuzheng and Hao, Jianye and Irwin, King},
  booktitle    = {Proceedings of the IEEE/CVF Conference on Computer Vision and Pattern Recognition (CVPR)},
  year         = {2024},
}

@inproceedings{bouvier2025DDAT,
  title        = {DDAT: Diffusion Policies Enforcing Dynamically Admissible Robot Trajectories},
  author       = {Bouvier, Jean‑Baptiste and Ryu, Kanghyun and Nagpal, Kartik and Liao, Qiayuan and Sreenath, Koushil and Mehr, Negar},
  booktitle    = {arXiv preprint arXiv:2502.15043},
  year         = {2025}
}

@inproceedings{liu2025embodiedTTA,
  title        = {Embodied Perception for Test‑Time Grasping Detection Adaptation with Knowledge Infusion},
  author       = {Liu, Jin and Xie, Jialong and Xiao, Leibing and Wang, Chaoqun and Zhou, Fengyu},
  booktitle    = {arXiv preprint arXiv:2504.04795},
  year         = {2025}
}

@inproceedings{imam2025TTL,
  title={Test‑Time Low Rank Adaptation via Confidence Maximization for Zero‑Shot Generalization (TTL)},
  author={Imam, Muntasir and others},
  booktitle={WACV},
  year={2025}
}

@inproceedings{chang2024PLUTO,
  title={Plug‑and‑Play Transformer Modules for Test‑Time Adaptation (PLUTO)},
  author={Chang, Xiangyu and Ahmed, Sk Miraj and Krishnamurthy, Srikanth V. and Guler, Basak and Swami, Ananthram and Roy‑Chowdhury, Amit K.},
  booktitle={arXiv preprint arXiv:2401.04130v3},
  year={2024}
}
}

\clearpage
\maketitlesupplementary

In the supplementary material, we report additional results and also provide more implementation details.

%-------------------------------------------
\section{Additional Results}

\subsection{Real-world Video Results}

To provide a more intuitive and comprehensive view of our real-world experiments, we have prepared an offline webpage containing video demonstrations of all tasks discussed in the main paper. This webpage allows for easy browsing the rollout results from our method \textbf{without requiring an internet connection}.

\paragraph{Accessing the Video Results via Web Browser.}
The video results are organized in a local HTML file included in the supplementary material zip package. To view them:
\begin{enumerate}
    \item Unzip the supplementary archive to a local folder.
    \item Open the \texttt{index.html} file located in the \texttt{AdaWorldPolicy\_Homepage/} \textbf{directory} by using any modern web browser (e.g., Chrome, Firefox).
    \item The page contains embedded players for all videos, which will play automatically.
\end{enumerate}

\paragraph{Content Organization.}
The webpage visualizes the performance of our method, \textbf{AdaWorldPolicy}, across four distinct manipulation tasks: \texttt{T1 - Sweep Coffee Beans}, \texttt{T2 - Long-horizon Pick-and-Place Eggs}, \texttt{T3 - Pour Water}, and \texttt{T4 - Wipe Whiteboard}. The video demonstrations contain two main parts:
\begin{itemize}
    \item \textbf{In-Domain Settings:} We first demonstrate the rollout results of our AdaWorldPolicy (without online adaptation) in the original training environments for all four tasks.
    \item \textbf{Domain Shift Settings:} We then showcase the effectiveness of our full method, AdaWorldPolicy (with AdaOL), under four challenging domain shift conditions for each task. These shifts include changes in tablecloths, the addition of distractors, changes in object instances, and random lighting variations.
\end{itemize}

All robotic arm operations shown in the videos are automatically generated through model inference and are accelerated by $10\times$ for efficient viewing.

\subsection{Visualization of Imagined Future Frames}

% 强调：经过我们的训练，可以作为 supervision，

\begin{figure*}[t]
        \centering
        \includegraphics[width=0.92\linewidth]{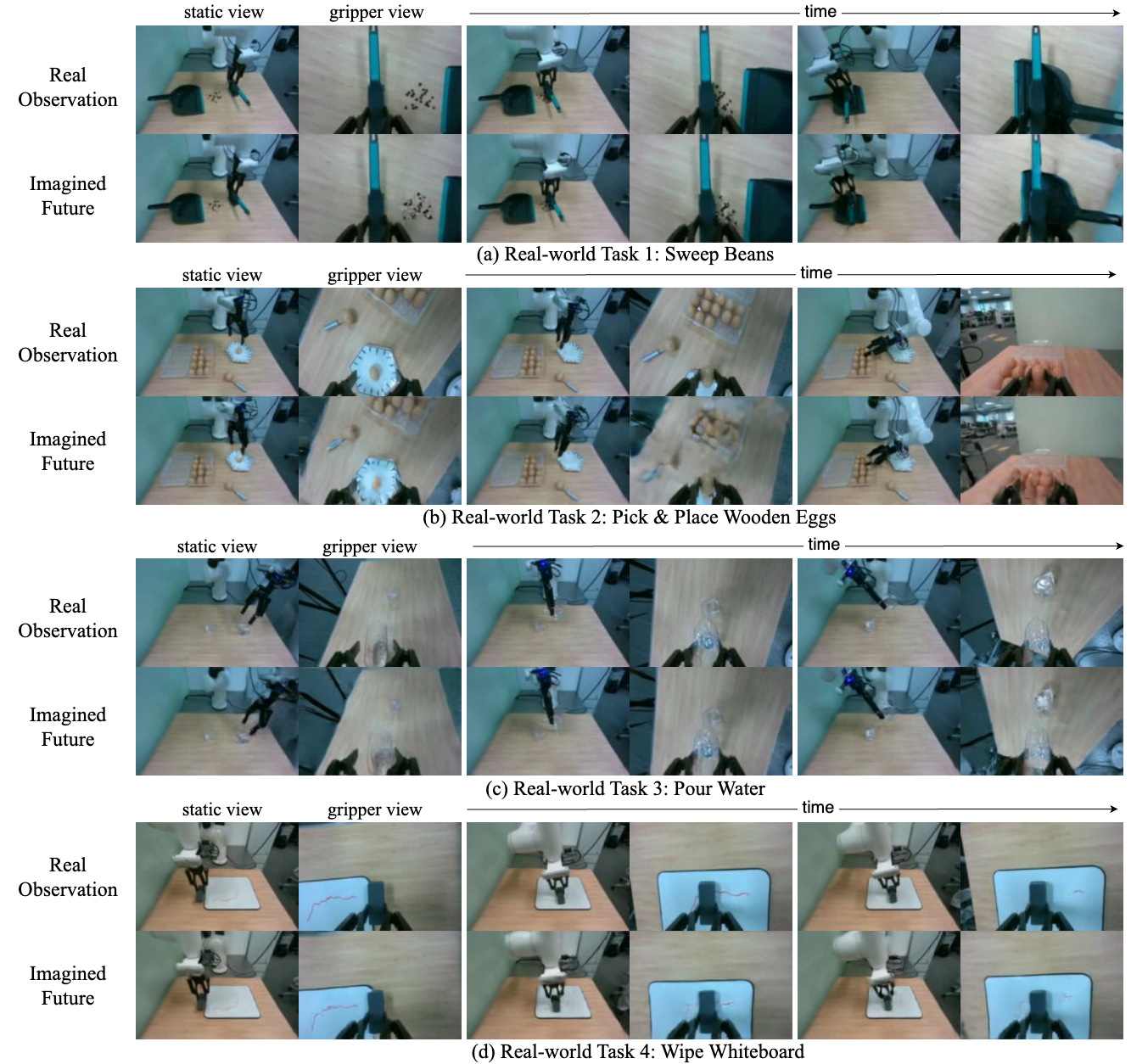} % Change the file name and format
        \caption{
        Visualization of imagined future frames in \textbf{real-world scenarios} generated by our AdaWorldPolicy. In each subfigures (a-d), we compare the imagined future frames (bottom row) with the actual real observations (top row) across various simulation and real-world tasks. In real-world scenarios (a-d), while the model generally captures the dynamics, some artifacts are observable due to the limitations of the base Cosmos-Predict2 model~\cite{cosmos2_ali2025world}, particularly in complex scenes with cluttered backgrounds or numerous objects, such as the egg manipulation task in (b).
        }
        \label{fig:imagination_vis_real}
        \vspace{-1em}
    \end{figure*}

To show that the world model can provide good supervision signals in our method, we provide a qualitative analysis of the future frames imagined by our AdaWorldPolicy across different tasks and domains. 
We compare the generated video predictions (Imagined Future) with the actual environmental observations (Real Observation) over time in both simulated benchmarks (see Fig.~\ref{fig:imagination_vis}) and real-world setup (see Fig.~\ref{fig:imagination_vis_real}). In simulation environments such as PushT~\cite{diffusionplanning_janner2022planning}, CALVIN~\cite{calvin_mees2022calvin}, and Libero10~\cite{libero_liu2024libero}(Fig.~\ref{fig:imagination_vis} a-c), our world model generates high-fidelity future frames that are highly consistent with the ground truth, effectively capturing the dynamics of the robot and objects. In real-world scenarios (Fig.~\ref{fig:imagination_vis_real}), the model successfully predicts the gripper's motion and key interactions. However, due to the inherent capacity limitations of the base video generation model (Cosmos-Predict2~\cite{cosmos2_ali2025world}), some visual artifacts and blurring are observable in complex real-world scenes, particularly in Task 2 (Fig.~\ref{fig:imagination_vis_real} (b)) where the background is cluttered and multiple small objects (eggs) exist. Despite observing these imperfect visual artifacts, the structural and semantic consistency remains sufficient for effective policy manipulation.

\begin{table}[t]
    \centering
    \caption{\textbf{Ablation study on sampling steps, adaptation, and fusion mechanisms.} We analyze the impact of inference sampling steps, the effectiveness of our AdaOL, and the choice of multi-modal fusion strategy. While reducing sampling steps slightly lowers performance, our method remains robust. Our AdaOL provides a consistent improvement. In addition, our MMSA module significantly outperforms the standard fusion baselines (Concatenation and Cross-Attention), validating its design. Note: AWP denotes our AdaWorldPolicy without using our online adaptive learning strategy (AdaOL). The symbol ``$\underline{~~~}$'' indicates that the result was reported in the main paper.}
    \label{tab:ablation_full}
    \resizebox{0.9\linewidth}{!}{%
        \begin{tabular}{lcc}
            \toprule
            \textbf{Configuration} & \textbf{\#Sampling Steps $\mathbb{\downarrow}$} & \textbf{Success Rate (\%) $\mathbb{\uparrow}$} \\
            \midrule
            \multirow{4}{*}{AWP} & 20 & 96.33 \\
                                       & 10 & \underline{95.53} \\
                                       & 5  & 94.67 \\
                                       & 2  & 94.00 \\
            \midrule
            + AdaOL                   & 10 & 96.05 \\
            \midrule
            MMSA  & 10 & 95.53 \\
            Concatenation   & 10 & 89.67 \\
            Cross-Attention & 10 & 91.21 \\
            \bottomrule
        \end{tabular}%
    }
\end{table}

\subsection{Ablation Studies on Libero10}

We further conduct comprehensive ablation studies on the Libero benchmark to validate key design choices of our AdaWorldPolicy, focusing on inference efficiency, adaptation strategies, and multi-modal fusion mechanisms. The results are summarized in Table~\ref{tab:ablation_full}.

\afterpage{
    \clearpage % 强制换页，开始放图
    \begin{figure*}[t]
        \centering
        % 调整高度为 0.95\textheight 以确保占满整页且不溢出
        \includegraphics[height=0.92\textheight]{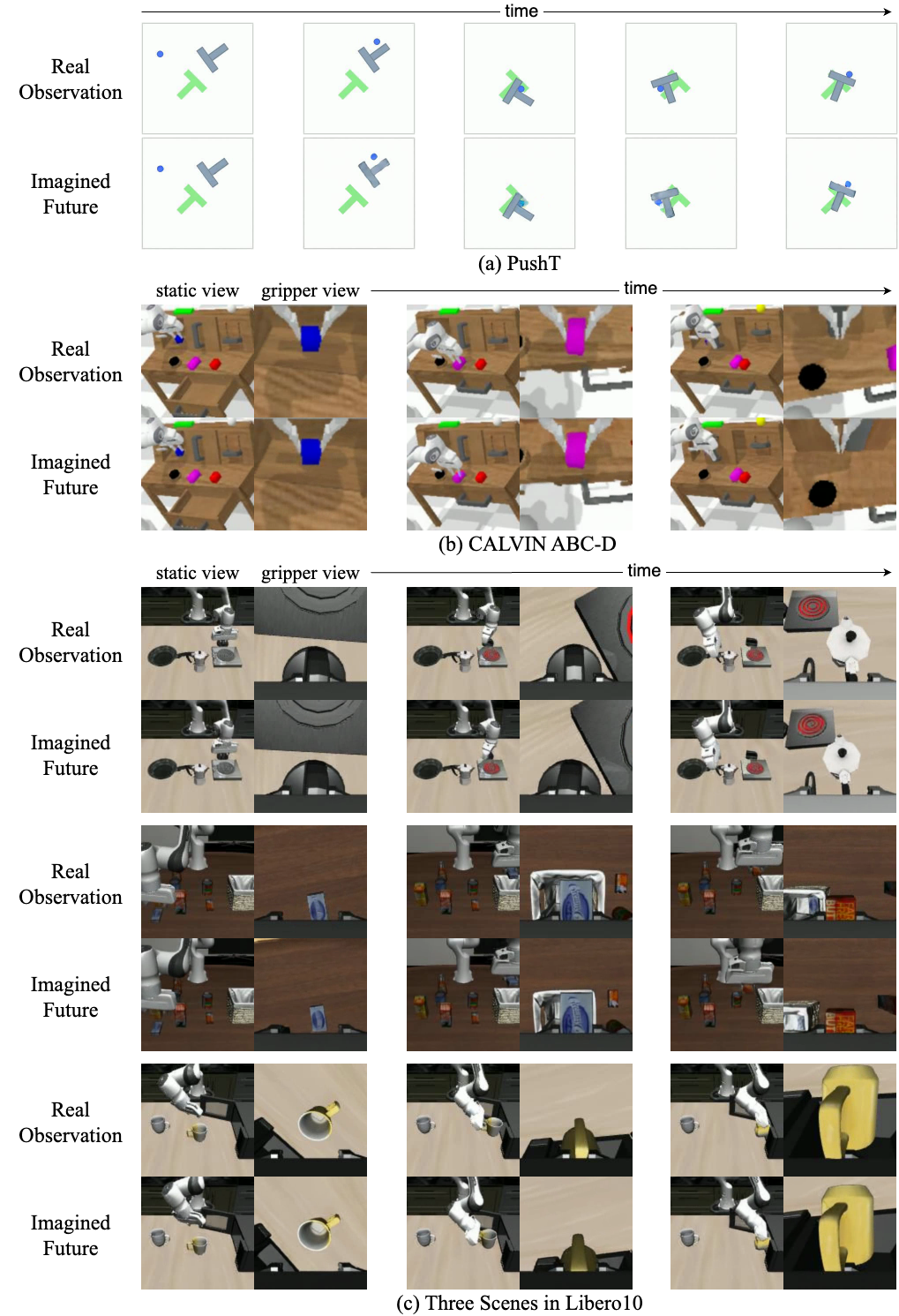} 
        \caption{
        Visualization of imagined future frames in \textbf{simulated benchmarks} generated by our AdaWorldPolicy.
        In each subfigure (a-c), we compare the imagined future frames (bottom row) with the actual real observations (top row). Across the PushT\cite{diffusionplanning_janner2022planning}, CALVIN\cite{calvin_mees2022calvin}, and Libero10~\cite{libero_liu2024libero} environments, the imagined frames demonstrate high consistency with the ground truth. While minor artifacts and slight deviations in agent position are present, the model effectively captures the core dynamics of the tasks.
        }
        \label{fig:imagination_vis}
    \end{figure*}
    \clearpage % 图片放完后，再次强制换页，恢复正文
}

\paragraph{Impact of Sampling Steps.}
We first investigate the trade-off between inference speed and performance by varying the number of denoising sampling steps. As shown in the first section of Table~\ref{tab:ablation_full}, the success rate exhibits a positive correlation with the number of steps. However, the policy generated from our method demonstrates remarkable robustness; even when the sampling steps are aggressively reduced to 2, the method maintains a competitive success rate of 94.00\%. This suggests that our diffusion-based policy learns a high-quality manifold that can be traversed efficiently.

\paragraph{Effectiveness of AdaOL.}
We further evaluate the contribution of our online adaptation mechanism. Enabling AdaOL (with 10 sampling steps) improves the success rate from 95.53\% to 96.05\%. This improvement indicates that AdaOL effectively mitigates subtle distribution shifts between the training data and the test environment, improving the policy's precision during execution.

\paragraph{Importance of MMSA Fusion.}
Finally, we analyze the efficacy of our Multi-Modal Self-Attention (MMSA) module by replacing it with standard fusion baselines while keeping other components fixed (using 10 sampling steps).
\begin{itemize}
    \item \textbf{MMSA $\rightarrow$ Concatenation:} Replacing MMSA with simple feature concatenation results in a significant performance drop to 89.67\%. This suggests that naive combination of visual and proprioceptive features is insufficient for capturing complex inter-modal dependencies.
    \item \textbf{MMSA $\rightarrow$ Cross-Attention:} Using a standard Cross-Attention mechanism yields a success rate of 91.21\%, which is still notably lower than our MMSA-based approach (95.53\%).
\end{itemize}
These results strongly validate the design of our MMSA, highlighting its superior ability to integrate multi-modal information for precise manipulation tasks than the conventional fusion approaches. This finding aligns with the observation drawn from our real-world ablation experiments presented in the main paper.

%-------------------------------------------
\section{Implementation Details}

\subsection{Network Details}

Our framework builds upon Cosmos-Predict2~\cite{cosmos2_ali2025world}, a world foundation model designed to generate future video frames conditioned on single-view video history and textual descriptions. While the model is available with both 2B and 14B parameters, we adopt the 2B version as our backbone to balance generation quality and computational efficiency under limited GPU resources. Although the original architecture supports action conditioning via Adaptive Layer Normalization (AdaLN)~\cite{adaln_huang2017arbitrary}, our empirical investigations revealed that this mechanism is sub-optimal for the specific task of action prediction. Consequently, we implement a separate action model branch integrated via our proposed Multi-Modal Self-Attention (MMSA)~\cite{sd3_esser2024scaling} mechanism. This decoupling design allows us to seamlessly utilize the pre-trained weights of the world model while maintaining full flexibility in the action model's architecture. Additionally, when extending the original single-view Cosmos-Predict2 to support multi-view inputs, we found that concatenating tokens from different views along the temporal dimension and sharing Rotary Positional Embeddings (RoPE)~\cite{rope_su2024roformer} lead to effective fine-tuning with stable convergence.

We tailor the scale of our models to suit the specific requirements of different experimental domains. In our real-world setup, which incorporates haptic feedback, we configure both the action model and the force predictor to have approximately 0.4B parameters each. Conversely, for simulation benchmarks where force data is unavailable, we reallocate the computational budget to enhance the policy's expressivity by increasing the size of the action model to 0.6B parameters. This adjustment allows the model to better capture complex manipulation behaviors in the absence of haptic cues, providing a performance boost with only a marginal increase in computational overhead.

\subsection{Data Processing}

In the learning-based robotic manipulation domain, it is well-established that larger training batch sizes generally correlate with improved policy performance and stability. However, within our AdaWorldPolicy framework, the video generation backbone (World Model) dominates GPU memory consumption. This creates a critical trade-off: while higher image resolutions are essential for the model to perceive fine-grained details necessary for precise manipulation (``seeing clearly''), they significantly increase the memory footprint, thereby limiting the maximum trainable batch size. To address this, we carefully tuned the hyperparameters for each benchmark, balancing training memory constraints, batch size, inference latency, and the required spatiotemporal resolution. The specific configurations for each environment are detailed in Table~\ref{tab:data_params}.

\begin{table}[t]
    \centering
    \caption{\textbf{Hyperparameter configurations for different benchmarks.} We adjust image resolution and temporal parameters to balance the trade-off between visual precision, memory consumption, and inference speed. Notably, the Real-world setting uses a sparse prediction strategy to minimize latency.}
    \label{tab:data_params}
    \resizebox{\linewidth}{!}{%
        \begin{tabular}{lcccc}
            \toprule
            \textbf{Benchmark} & \textbf{Image Size} & \textbf{History Length} & \textbf{Action Horizon} & \textbf{\#Imagined Frames} \\
            \midrule
            Libero10   & $128 \times 128$ & 5 & 20 & 20 \\
            PushT      & $256 \times 256$ & 5 & 20 & 20 \\
            CALVIN     & $192 \times 192$ & 1 & 12 & 12 \\
            Real-world & $112 \times 160$ & 1 & 32 & 4  \\
            \bottomrule
        \end{tabular}%
    }
\end{table}

It is worth noting that for the Real-world setting, the number of imagined frames (4) is significantly lower than the action horizon (32), unlike in the simulation benchmarks. This design choice was made to minimize inference latency. We implemented a \textit{skipped frame prediction} strategy, where the model predicts future frames at fixed intervals rather than generating a dense sequence. This approach allows the model to cover the full temporal span of the action horizon while drastically reducing the computational cost associated with video generation.

In our real-world experiments, we also observed that the preprocessing of force data requires distinct handling process compared to image or relative action modalities. Unlike pixel values or relative end-effector poses, force measurements typically lack strict theoretical upper or lower bounds and can exhibit significant variance or spikes. Consequently, standard normalization techniques (e.g., min-max scaling based on absolute extremes) can be unstable or lead to compressed data distributions. We empirically found that employing \textbf{quantile-based normalization}—specifically, scaling data based on statistical percentiles (i.e., the $1^{st}$ and $99^{th}$ percentiles)—provides a robust mapping. This strategy effectively mitigates the impact of outliers and resulted in the optimal balance of convergence speed and final policy performance.

\subsection{Real-world Evaluation Protocol}

\paragraph{Data Collection and Training.}
For each of the four real-world manipulation tasks, we collected 150 expert demonstrations in the in-domain environment via teleoperation. These datasets were utilized to train the policy using standard offline behavior cloning. Once trained, the models were evaluated directly in both the original in-domain setting and four distinct domain-shift scenarios to assess robustness and generalization capability.

\paragraph{Evaluation Metrics and Baselines.}
For every model configuration, we conduct 30 evaluation trials per task under each domain distribution. A maximum limit of 1500 execution steps is enforced for all tasks. To ensure a fair comparison, we enhanced the standard Diffusion Policy baseline to incorporate haptic feedback. Specifically, 6-dimensional force data is injected into the model by concatenating it with image observation features and processing it via cross-attention within the Transformer architecture. We refer to this haptic-enabled baseline as \textbf{DP-Force}.

\paragraph{Success Criteria.}
The specific success conditions for the four tasks are defined as follows:
\begin{itemize}
    \item \texttt{T1 - Sweep Beans:} The task is considered successful if no more than three coffee beans (or corn kernels) remain on the table surface.
    \item \texttt{T2 - Pick and Place Eggs:} The egg must be transported without touching the table surface during the whole trajectory and must be placed securely into an empty slot in the carton.
    \item \texttt{T3 - Pour Water:} The final volume of water in the target glass must exceed 90\% of the initial water volume in the source cup.
    \item \texttt{T4 - Wipe Whiteboard:} The length of any remaining marker trace on the whiteboard must not exceed 3 cm.
\end{itemize}

\paragraph{AdaOL Testing Protocol.}
To rigorously evaluate the effectiveness of our online adaptation mechanism, we adopt a specific protocol for the AdaOL experiments. For each task and domain combination, the model is reset to its pre-trained state before the first rollout. The evaluation proceeds in two sequential phases. First, during the adaptation phase (Trials 1-15), the model performs the task while continuously updating its parameters online via test-time adaptation, with the outcomes of these trials recorded. Subsequently, during the frozen phase (Trials 16-30), the model weights are frozen, and the remaining 15 trials are executed without further updates to evaluate the stability of the adapted policy. The final reported success rate is the overall performance from all 30 trials.

\section{Limitation and Future Work}

Despite our AdaWorldPolicy has achieved promising results, we point out several limitations that may lead to future research directions.

First, while our backbone world model, Cosmos-Predict2~\cite{cosmos2_ali2025world}, is pre-trained on millions of hours of video data and represents the state-of-the-art embodied world modeling capability, its generalization capabilities remain imperfect under significant domain shifts. In our experiments, particularly within real-world settings, we observed that the quality of predicted future frames degrades when the environment undergoes drastic changes, such as continuous and varying lighting conditions. This suggests that this world model still lacks sufficient robustness to handle extreme out-of-distribution scenarios. However, it is important to note that our primary objective is effective policy execution rather than photorealistic video generation. Our results demonstrate that our method can generate reasonable policy even when the predicted visual future contains minor artifacts. We anticipate that as stronger and more generalizable world models emerge, integrating them into our framework will naturally alleviate these visual distortions and further enhance execution performance.

Second, regarding our online adaptation mechanism, we employed a fixed set of hyperparameters for AdaOL across all domain shift experiments to demonstrate the method's general applicability. We did not perform task-specific or environment-specific fine-tuning of the adaptation parameters. Future work could explore adaptive or meta-learning approaches to automatically tune AdaOL's hyperparameters during deployment, which may potentially lead to further performance improvements.

% \clearpage
% {
%     \small
%     \bibliographystyle{ieeenat_fullname}
%     \bibliography{main}
% }

% {
%     \small
%     \bibliographystyle{ieeenat_fullname}
%     \bibliography{main}
% }

\end{document}